\documentclass{article}
\usepackage{multirow}

\PassOptionsToPackage{numbers, compress}{natbib}

\usepackage[preprint]{neurips_2025}

\usepackage[utf8]{inputenc} 
\usepackage[T1]{fontenc}    
\usepackage{hyperref}       
\usepackage{url}            
\usepackage{booktabs}       
\usepackage{wrapfig}        
\usepackage{amsfonts}       
\usepackage{graphicx}       
\usepackage{nicefrac}       
\usepackage{microtype}      
\usepackage{xcolor}         
\usepackage{xspace}
\usepackage{amsmath}
\usepackage{comment}
\usepackage[fixed]{fontawesome5}
\usepackage{amssymb}
\usepackage{xcolor}         
\usepackage{subcaption}
\usepackage[most]{tcolorbox}  
\usepackage{diagbox}
\usepackage{bbm}
\usepackage{makecell}

\definecolor{pastcolor}{HTML}{009392}
\definecolor{presentcolor}{HTML}{E2E29C}
\definecolor{futurehypocolor}{HTML}{EEB379}
\definecolor{futurefactcolor}{HTML}{CF597E}

\tcbset{on line, boxsep=1pt, left=1pt,right=1pt,top=0pt,bottom=0pt, colback=pastcolor}

\newcommand{\dataset}{AnytimeVQA-1K\xspace}
\newcommand{\model}{AViLA\xspace}

\newcommand{\qea}{Query-Evidence Asynchrony\xspace}

\newcommand{\memicon}[0]{{\small{\faBrain}}}
\newcommand{\evidenceicon}[0]{{\small{\faGlasses}}}

\newcommand{\triggericon}[0]{{\small{\faLightbulb}}}

\newcommand{\tanveer}[1]{\textcolor{red}{[{\bf Tanveer:} #1]}}

\newcommand{\pastq}{\tcbox[colback=pastcolor!50, boxrule=0pt, arc=3pt]{Past}\xspace}
\newcommand{\presentq}{\tcbox[colback=presentcolor!50, boxrule=0pt, arc=3pt]{Present}\xspace}
\newcommand{\futurehypoq}{\tcbox[colback=futurehypocolor!50, boxrule=0pt, arc=3pt]{Future-Prediction}\xspace}
\newcommand{\futurefactq}{\tcbox[colback=futurefactcolor!50, boxrule=0pt, arc=3pt]{Future-Observation}\xspace}

\makeatletter
\DeclareRobustCommand\onedot{\futurelet\@let@token\@onedot}
\def\@onedot{\ifx\@let@token.\else.\null\fi\xspace}

\def\eg{\emph{e.g}\onedot}

\def\etc{\emph{etc}\onedot}

\makeatother

\title{\model: Asynchronous Vision-Language Agent for Streaming Multimodal Data Interaction}

\author{%
  Gengyuan Zhang$^{1,2}$\thanks{Equal contribution} \quad
  Tanveer Hannan$^{1,2}$\footnotemark[1] \quad
  Hermine Kleiner$^{1}$\thanks{Equal second contribution} \quad
  Beste Aydemir$^{1}$\footnotemark[2] \\
  \textbf{Xinyu Xie}$^1$ \quad
  \textbf{Jian Lan}$^{1,2}$ \quad
  \textbf{Thomas Seidl}$^{1,2}$ \quad
  \textbf{Volker Tresp}$^{1,2}$ \quad
  \textbf{Jindong Gu}$^3$ \\
  $^1$LMU Munich \quad
   $^2$Munich Center for Machine Learning \quad
   $^3$University of Oxford
   \\
  \texttt{\{zhang, hannan\}@dbs.ifi.lmu.de}
}

\begin{document}

\maketitle

\begin{abstract}
An ideal vision-language agent serves as a bridge between the human users and their surrounding physical world in real-world applications like autonomous driving and embodied agents, and proactively provides accurate and timely responses given user intents.
An intriguing challenge arises when agents interact with the world as a dynamic data stream and ad-hoc queries from users: supporting knowledge for queries, namely evidence, usually appears asynchronously with the arrival time of queries, and agents need to ground their responses in historical data, present observations, and even future streams.
We frame this challenge as \qea, where user queries and their supporting evidence typically arrive asynchronously in the streaming setting. 
This setting requires not only strong reasoning capabilities but also the ability to retain past observations and respond to queries with temporal awareness.
In this paper, we introduce a diagnostic benchmark that evaluates Multimodal Large Language Models (MLLMs) on their ability to handle interaction with streaming data.
Further, we present \model, Asynchronous Video-Language Agent for streaming data interaction that can handle ad-hoc queries and give time-aware responses. For this purpose, \model consists of three key modules: \textit{comprehensive memory retention}, \textit{evidence identification}, and \textit{evidence-grounded trigger}, that are designed to maintain a general-purpose memory and respond readily and timely to queries.
Our experiments show that existing models often fail to respond at appropriate times, while \model significantly improves both accuracy and temporal awareness.
Our code and dataset will be publicly available.

\end{abstract}

\section{Introduction}\label{sec:intro}

\begin{figure}[htbp]
    \centering
    \includegraphics[width=1\linewidth]{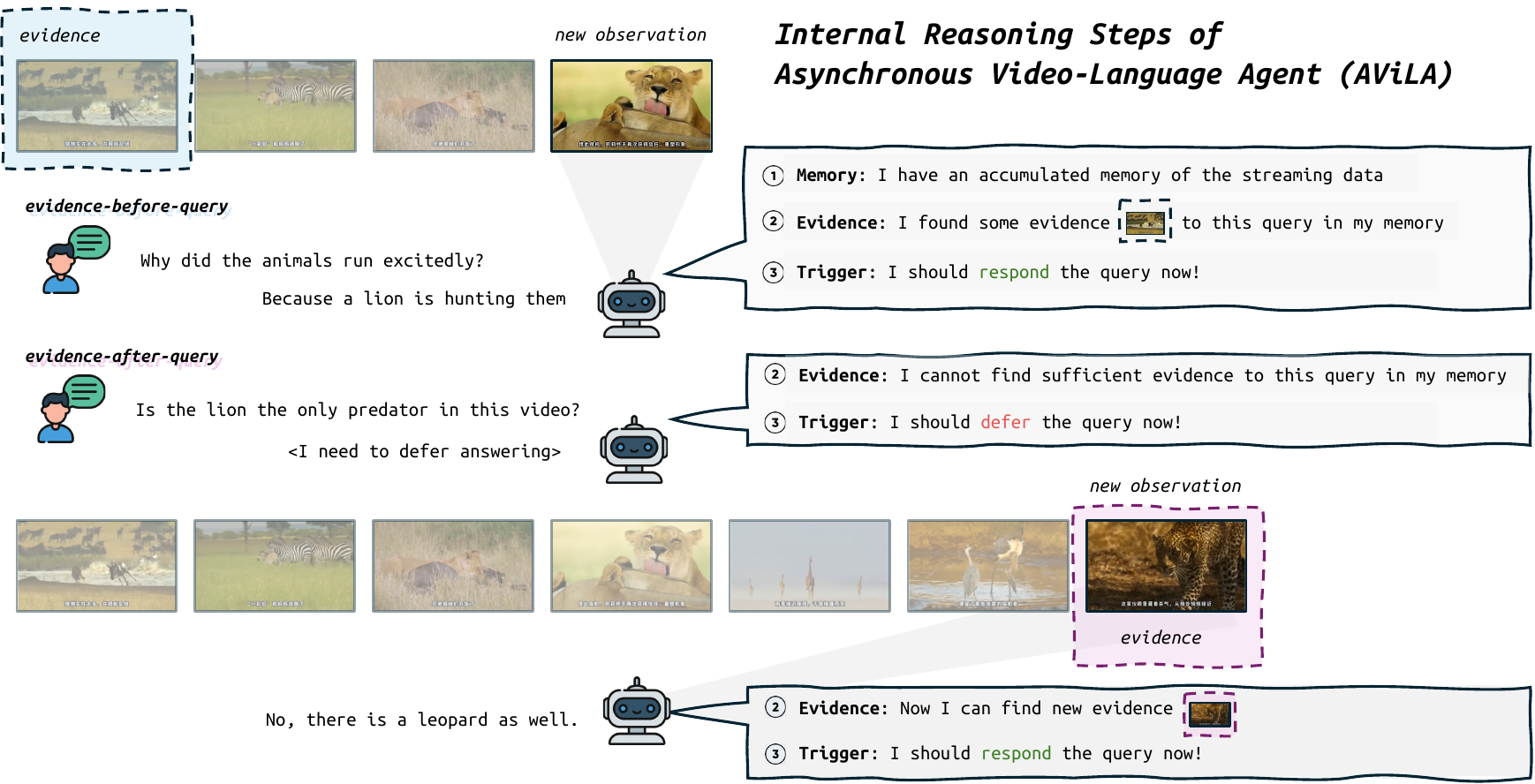}
    \caption{When interacting with streaming video data, user queries may arrive at any time, while supporting evidence may have already appeared or may not yet be observable. We present \model, which maintains a query-agnostic memory of past visual data and assess whether sufficient evidence exists to respond to the query. For each query, \model performs three steps: (1) It checks its accumulated memory, (2) searches for relevant evidence, and (3) invokes a trigger mechanism to decide whether to respond or defer the query.}
    \label{fig:teaser}
\end{figure}

Real-world applications, such as autonomous vehicles~\cite{stappen2023integrating}, embodied robotics~\cite{driess2023palm,mu2023embodiedgpt}, and web copilots~\cite{zheng2024gpt}, heavily rely on streaming multimodal data like videos and represent a pivotal step towards Artificial General Intelligence (AGI). 
Streaming videos are a key interface between the physical world, the user, and the agent.
While recent advances in Vision-Language Agents~\cite{bai2023qwen, bain2021frozen, chen2024sharegpt4video, Chen_2024_CVPR, hong2024cogagent,tang2025video, weng2024longvlm} have shown strong reasoning capability on offline and holistic videos~\cite{lei2021less, li2023videochat, maaz2023video} across tasks like question answering~\cite{chen2023minigpt, fan2024videoagent, qian2024streaming, wang2024videoagent} and summarization~\cite{argaw2024scaling, hua2025v2xum, guo2025vtg, wu2024videollm}, their applicability to streaming scenarios remains limited.
In contrast to offline settings, interacting with streaming video data introduces the challenges of a temporal mismatch between the moment when an ad-hoc query is raised and the time when the relevant evidence emerges in the stream. Imagine you're watching a live soccer match. During the match, you ask: ``\texttt{Did anyone notice if Smith was limping after that tackle in the first half?}'' to recall past events or ``\texttt{I am taking a nap, keep watching and tell me when someone scores.}'' Such queries demand searching for evidence in the past observations or waiting for evidence to appear.

We refer to this as the \textit{\qea} problem in streaming data interaction, which remains underexplored in both benchmarks and modeling.
In many real-world applications, such as embodied agents that assist in household tasks or autonomous vehicle agents, queries and evidence come asynchronously.
For example, (1) ``\texttt{Which object did the person pick up before exiting the room?}'' requires retrieving past observations (\textit{evidence-before-query}); (2) ``\texttt{What is the person doing right now?}'' depends on current perception (\textit{evidence-during-query}); and (3)``\texttt{Will the person complete the assembly task correctly?}'' involves waiting for future evidence to unfold (\textit{evidence-after-query}).
These asynchronous interactions highlight the need for agents to reason over evolving observations and decide when sufficient evidence has accumulated to provide a grounded and timely response.

We hypothesize that \qea arises from three contributing factors: 
(1) On the \textbf{data} side, unbounded input streams grow continuously over time, necessitating effective memory mechanisms to process and retain relevant observations;
(2) On the \textbf{user} side, ad-hoc queries may arrive unpredictably and without prior specification, requiring the model to maintain general-purpose, query-agnostic representations of past and present observations;
(3) On the \textbf{model} side, generated responses must be time-aware: that is, the model should respond \textit{readily} only when sufficient evidence is available, and \textit{timely} to minimize unnecessary delays once that evidence is observed.

To facilitate a systematic study of the \qea problem, we first introduce a high-quality diagnostic benchmark, \dataset (Sec.~\ref{subsec:benchmark}), containing 1,000 question-answer pairs spanning 189 videos (up to 90 minutes) long from diverse domains.
Unlike existing similar benchmarks and datasets~\cite{qian2024streaming, li2025ovobenchfarvideollmsrealworld,yang2025svbench}, \dataset is carefully constructed to involve asynchronous queries for streaming data by distinguishing itself in: (1) we employ a general-purpose query design such that models should not have prior knowledge of the query type and temporality (unlike ~\cite{li2025ovobenchfarvideollmsrealworld}); (2) we attempt to introduce an asynchrony-critical question design that demands more challenging contextual and temporal understanding from models and exclude questions that can be answered by current observations unlike ~\cite{qian2024streaming} (Sec.~\ref{subsec:benchmark}).
Beyond using accuracy as the sole metric, our evaluation also emphasizes temporal awareness of responses, requiring models to respond not only correctly, but also timely, and at the appropriate moment. This benchmark enables a nuanced empirical evaluation of state-of-the-art MLLMs in handling asynchronous queries.

Preceding and concurrent methods fail to provide a generic solution to handling asynchronous queries in streaming data (explained in Sec.~\ref{sec:relate})
To confront these challenges, we propose \textbf{\model} (the \textbf{A}synchronous \textbf{Vi}deo-\textbf{L}anguage \textbf{A}gent), a framework equipped to operate under \qea. 
In streaming environments, agents must contend with incomplete, evolving contexts while maintaining the ability to respond to ad-hoc queries. 
Thus, as illustrated in Fig.~\ref{fig:teaser}, AViLA is built upon three essential modules: (1) \textit{comprehensive memory retention}, which retains temporally evolving visual input in a question-agnostic manner to confront arbitrary future queries; (2) \textit{query-guided evidence identification}, which retrieves query-relevant evidence from the comprehensive memory to determine if the query is answerable; and (3) an \textit{evidence-grounded trigger}, which determines whether to respond or defer the query given the identified evidence in the last step. These components enable temporal-aware reasoning over streaming video data.

Our contributions are summarized as follows:
\begin{enumerate}
\setlength\itemsep{-0.3em}
\item  We introduce \qea as a challenge in handling streaming multimodal data interactions, when ad-hoc user queries and their supporting evidence arrive asynchronously;
\item We develop a comprehensive diagnostic benchmark dataset (\dataset) that facilitates systematic evaluation, featuring diverse query categories and complexity in this challenge;
\item We propose \model, an agentic framework to combat this challenge. Our extensive experiments demonstrate its superior performance to provide accurate and time-aware responses to user queries.
\end{enumerate}

\section{Related Work}\label{sec:relate}
\vspace{0.2cm}
\textbf{Video Language Models}
Recent advances in Large Language Models (LLMs)~\cite{bi2024deepseek, team2024gemma, touvron22023llama} and Multimodal Large Language Models (MLLMs)~\cite{alayrac2022flamingo, li2023blip, zhu2023minigpt} have sparked growing interest in applying these models to holistic video understanding tasks~\cite{cheng2024videollama, liu2023visual, zhang2023video, wang2024qwen2}.
Extending these capabilities to streaming videos has become an increasingly active area of research.
A straightforward approach involves adapting general-purpose MLLMs to streaming scenarios. StreamingBench~\cite{lin2024streamingbenchassessinggapmllms} explores this direction by benchmarking both proprietary models (\eg, GPT-4o~\cite{hurst2024gpt}, Gemini~\cite{team2023gemini}, Claude~\cite{claude3report2024}) and open-source systems (\eg, Video-LLaMA2~\cite{cheng2024videollama}, MiniCPM~\cite{hu2024minicpm, yao2024minicpm}, InternVL-V2~\cite{chen2024internvl}) on streaming video tasks.

Meanwhile, a few models have been proposed and specialized for streaming video understanding. Flash-VStream~\cite{zhang2024flashvstreammemorybasedrealtimeunderstanding} introduces a custom memory-augmented architecture that processes each frame with a visual encoder and passes accumulated memory to an LLM when a query arises. VideoLLM-Online~\cite{Chen_2024_CVPR} employs a dialogue-based framework with a streaming-aware training objective, enabling live multi-turn interaction throughout the stream.
Dispider~\cite{qian2025dispider} is the most relevant method to ours, yet it is only focused on the trigger design to determine evidence efficiency but ignores the necessity and challenges of comprehensive memory when handling ad-hoc user queries.

\vspace{0.2cm}
\textbf{Vision-Language Agents}
Agentic models have emerged as a promising paradigm for addressing a wide range of video understanding tasks, leveraging LLMs and MLLMs. We refer to these works collectively as \textit{Vision-Language Agents} that integrate MLLMs with strategies such as external tool use, episodic memory, and symbolic reasoning to perform tasks including video question answering~\cite{gao2023mist, liao2024align, min2024morevqa, wang2024videoagent,zang2023discovering}, summarization~\cite{li2024unsupervised, liu2023video, sul2023mr, zengtimesuite}, and localization~\cite{barrios2023localizing, chen2024learning,  rahman2024deeplocalization,  Zhang_2024_WACV}.
VideoAgent~\cite{wang2024videoagent} pioneered this line of work by incorporating MLLMs with external tools to solve video understanding tasks. Concurrent and subsequent works~\cite{fan2024videoagent, kahatapitiya2025language, wang2024videollamblongcontextvideounderstanding} extend this direction by treating videos as contexts and fully leveraging the reasoning capabilities of LLMs over temporally dense content.
In parallel, systems like ChatVideo~\cite{wang2023chatvideotrackletcentricmultimodalversatile}, DoraemonGPT~\cite{yang2024doraemongptunderstandingdynamicscenes}, and OmAgent~\cite{zhang2024omagentmultimodalagentframework} convert video streams into symbolic or database-like structures, enabling compositional reasoning over structured representations.
A recurring design component across most of these systems is using memory architectures to support long-horizon or egocentric video comprehension. VideoAgent~\cite{fan2024videoagentmemoryaugmentedmultimodalagent}, MM-VID~\cite{lin2023mmvidadvancingvideounderstanding}, and LifelongMemory~\cite{wang2024lifelongmemoryleveragingllmsanswering} incorporate memory-augmented reasoning to track evolving video content and support context-dependent responses.
To bridge the gap between visual inputs and language reasoning in MLLMs, many approaches convert videos into dense textual narrations or captions~\cite{kahatapitiya2025language, wang2024videoagent,zhang2024mm,yang2024vca} as representations that serve as an anchor for downstream tasks.


\section{\qea in Streaming Multimodal Data Interaction}
\subsection{Preliminaries}
Let the input streaming data be represented as a sequence of frames, each regarded as new observation arriving over discrete time steps:
$\mathcal{V} = \{v_1, v_2, v_3, \dots \}$, where $v_t$ denotes the visual observation at discrete time step $t$ and \( T \) denotes the total stream length.
A query \( q_i \in \mathcal{Q} \) may arrive at any time step \( \tau(q_i) \in [1, T] \), and may reference evidence that occurred in the past, present, or future relative to \( \tau(q_i) \).
Evidence $\mathcal{E}(q_i)$ for the query $q_i$ refers to the necessary information required to answer the question, which may reside in the past stream, the present observation, or future streams, and is independent of $\tau(q_i)$.
This temporal mismatch leads to \qea (Sec.~\ref{sec:eqa}).

At time \( t \), the agent perceives a new frame \( v_t \) and maintains a memory bank \( \mathcal M_t \) to retain the observed information:
\[
\mathcal M_t = \mathrm{Retention}(\mathcal M_{t-1}, p_t), \quad p_t = \mathrm{Ingestion}(v_t)
\]
where \( \mathrm{Retention} \) is a generic memory retention function, and \( \mathrm{Ingestion} \) is the perception module.

At each time step $t$, the agent must determine an action from $a_i \in \{\texttt{Respond}, \texttt{Defer}\}$ based on the current memory $\mathcal{M}_{t}$ for each query \( q_i \) if $\tau(q_i) \leq t$. However, it is impractical to feed the whole memory $\mathcal{M}$ to the agent, so we consider only necessary information, namely evidence, should be identified from memory and utilized.
\[
a_i = \mathrm{Trigger}(q_i, \mathcal{E}_t(q_i), p_{t}), \quad \mathcal{E}_t(q_i) = \mathrm{Identification}(\mathcal{M}_t, q_i)
\]
If the agent determines to respond at timestep $t_i^{respond}$, it should further output an answer. Otherwise, it defers answering to a future time \( t' > \tau(q_i) \), until it assumes sufficient evidence is available.

\subsection{\qea}\label{sec:eqa}
We define \qea as the temporal mismatch between the time a user query arrives and the time at which sufficient supporting evidence becomes available in the data stream.

\vspace{0.2cm}
\textbf{Evidence and Query}
According to the temporal relationship between query and evidence on the streaming timeline, we categorize it into three cases: \textit{evidence-before-query}, \textit{evidence-during-query}, and \textit{evidence-after-query}. 
Evidence-before-query refers to cases where the relevant evidence has already occurred before the query time. In this setting, the agent must retrieve information from its memory, such as memorization queries like ``\texttt{What has the man done to irritate his wife?}''
Evidence-during-query corresponds to questions about events unfolding in real time, such as narration queries like ``\texttt{What is happening in the current frame?}'' that require immediate observation and ingestion.
Evidence-after-query involves queries that cannot be answered immediately due to insufficient information at the time of asking. These require the agent to defer its response until the necessary evidence becomes available in future observations, for instance, queries like ``\texttt{Who will enter the room next?}'' or ``\texttt{Does the person pick up the object later?}''

\vspace{0.2cm}
\textbf{Agent Principles} 
In light of the three causes underlying \qea (Sec.~\ref{sec:intro}), we argue that a competent agent must satisfy the following three principles:

\begin{itemize}

    \item \textit{Comprehensiveness}: Since an arbitrary \( q_i \) is unknown in advance, the agent must maintain a comprehensive memory \( M_t \) that retains observations to support any possible queries.

    \item \textit{Readiness}: The agent should respond only when sufficient evidence has been accumulated in memory. Premature responses based on partial evidence should be avoided.

    \item \textit{Timeliness}: Once sufficient evidence becomes available, the agent should respond without unnecessary delay; passive waiting until the end of the stream is undesirable.
\end{itemize}


\subsection{Diagnostic benchmark: \dataset}
\label{subsec:benchmark}
We present a new diagnostic benchmark, \dataset. \dataset comprises 1,000 human-annotated high-quality questions and 189 videos sourced from four existing datasets~\cite{Ego4D, MovieChat, CrossTask, Video-MME}. Each video is accompanied by multiple queries at different timestamps. Videos selected range from 6 minutes up to one and a half hours, with an average length of around 22 minutes, and span a broad spectrum of domains - including everyday life, movies and TV, sports, documentaries, and news reports - offering realistic settings for streaming video interaction. This diversity allows us to rigorously evaluate a model’s generalization and temporal reasoning capabilities under \qea. The annotation process was extensive, with annotators collectively spending approximately 400 hours to produce high-quality, time-stamped queries. We elaborate on further dataset details in the Appendix (Sec.~\ref{sec:more-dataset}).

\begin{figure}[!h]
    \centering
        \begin{subfigure}{0.4\linewidth}
        \centering
        \includegraphics[width=\linewidth]{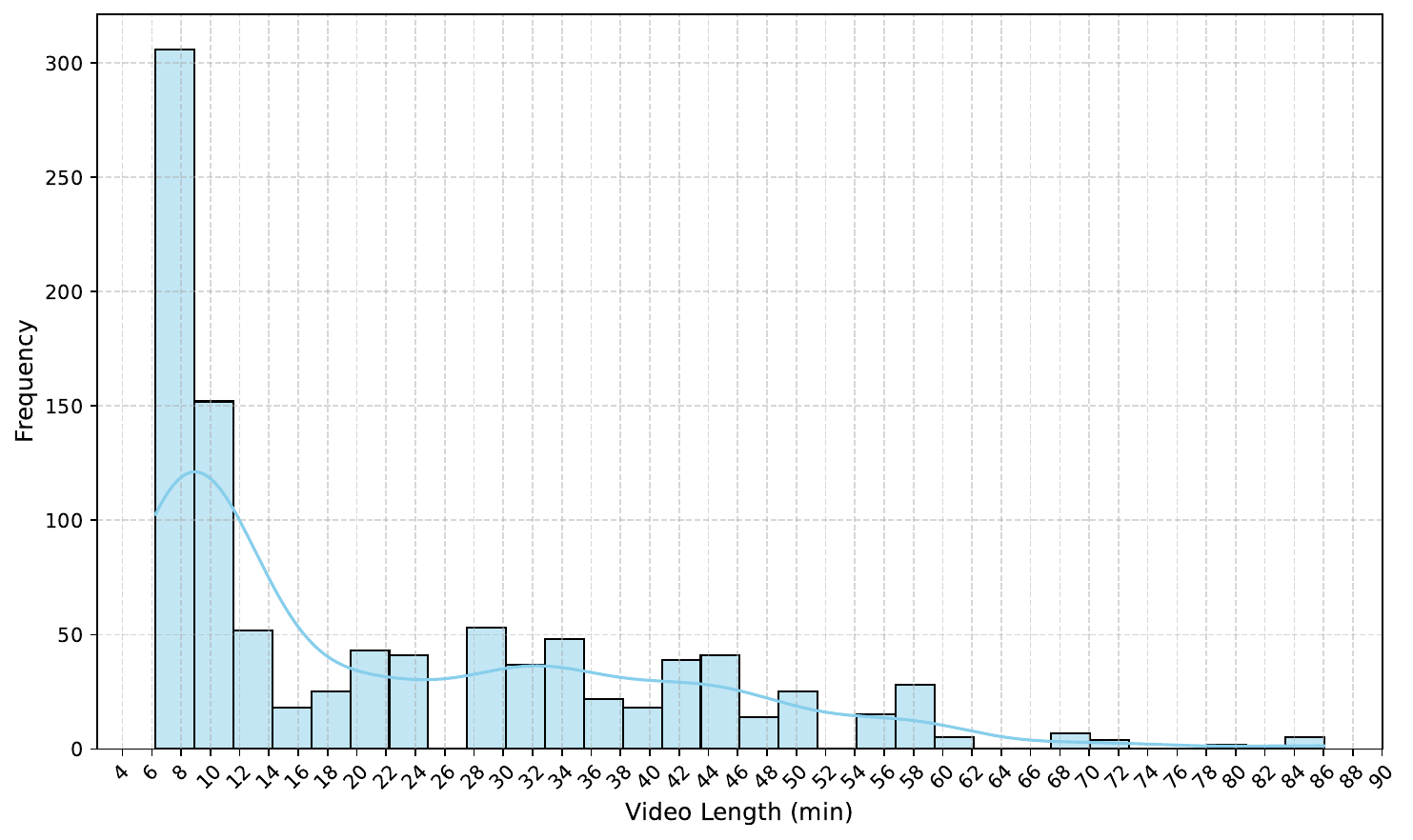}
        \caption{Video lengths in \dataset.}
        \label{fig:video_len}
    \end{subfigure}
    \hspace{0.4cm}
    \begin{subfigure}{0.4\linewidth}
        \centering
        \includegraphics[width=\linewidth,trim=0 25 0 15, clip]{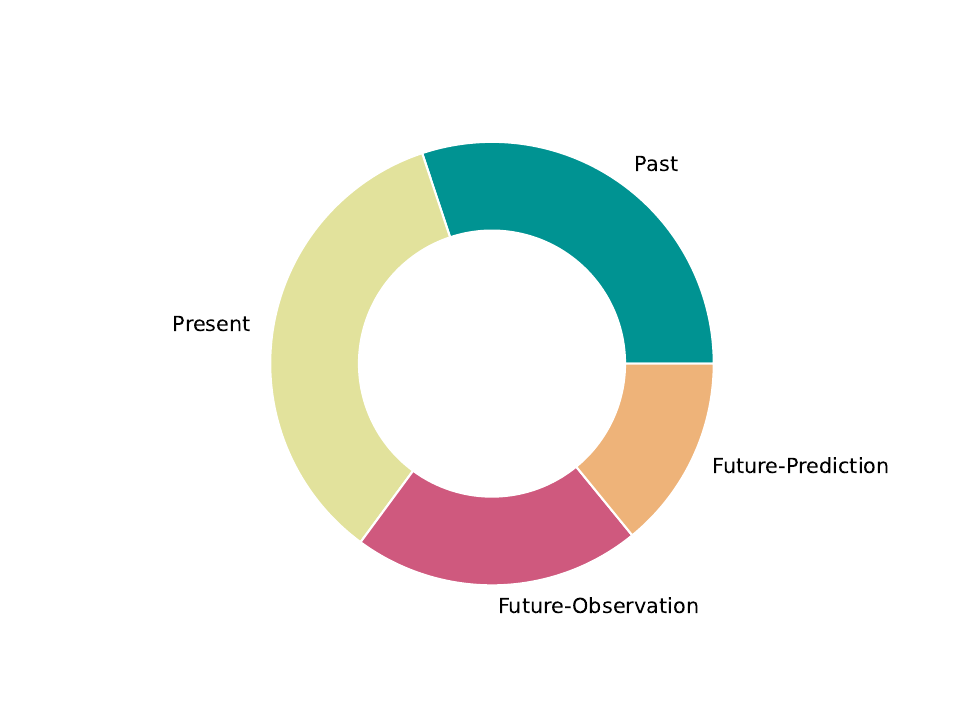}
        \caption{Query distribution in \dataset.}
        \label{fig:question-cat}
    \end{subfigure}

    \vspace{0.1cm}
    \begin{subfigure}{1.\linewidth}
        \centering
        \includegraphics[width=\linewidth]{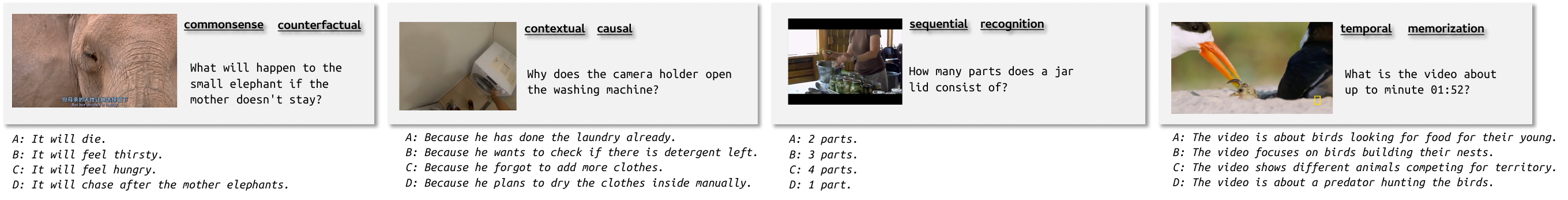}
        \caption{Examples of \dataset.}
        \label{fig:question-example}
    \end{subfigure}
    \caption{We present \dataset as a diagnostic benchmark to study \qea in streaming multimodal data.}
    \label{fig:question-types}
    

\end{figure}

\vspace{0.2cm}
\textbf{Query Temporality}
To better understand how \qea affects model behavior, we categorize questions into three categories based on their temporal relation to the required evidence: \pastq, \presentq, \futurehypoq, \futurefactq.

\pastq queries are retrospective questions requiring the model to answer based on evidence shown \textit{before} the query. The agent must recall and reason over previously observed events without access to the earlier segments of video episodes.
\presentq queries are questions about events occurring at the time the query is issued, which demands temporal and contextual knowledge \textit{before} and current observation \textit{during} the query.
Future queries refer to events that have not yet occurred at the time of questioning or do not factually happen in the video episode. We distinguish between \futurehypoq queries, which are forward-looking and predictive questions based on the current knowledge (\textit{evidence-before/during-query}), such as planning and forecasting tasks, and \futurefactq queries, which rely on perceiving events or observations that will be verified at a future time (\textit{evidence-after-query}).

\emph{Notably, models should not know the category of a query} and must determine if the evidence is sufficient at the current timestep, necessitating the importance of time-aware responses.

\vspace{0.2cm}
\textbf{Evaluation}
As highlighted by the challenges inherent in \qea, we propose evaluating model responses from two key perspectives: (1) \textit{accuracy}, and (2) \textit{temporal awareness}.
We follow a standard multiple-choice question answering (MCQ) setup and report the average accuracy over \dataset.

To evaluate the temporal awareness, we first define a temporal offset $\delta_i$ for response time $t_i^{\text{respond}}$ for each question $q_i$, relative to the \textit{proposed answering window} $[t_i^{\text{start}}, t_i^{\text{end}}]$ annotated in the dataset. $\delta_i$ is calculated as follows:
\[
\delta_i = \begin{cases}
t_i^{\text{respond}} - t_i^{\text{start}} & \text{if } t_i^{\text{respond}} < t_i^{\text{start}} \\
0 & \text{if } t_i^{\text{start}} \leq t_c \leq t_i^{\text{end}} \\
t_i^{\text{respond}} - t_i^{end} & \text{if } t_i^{\text{respond}} > t_i^{\text{end}}
\end{cases}
\]

A negative $\delta_i$ indicates that the response is given before sufficient evidence is available, while a positive $\delta_i$ means the response is delayed beyond the proposed answering window. 
The temporal offset captures whether the model makes a premature guess without sufficient evidence or adopts a ``lazy'' strategy by blindly deferring its response until the last possible moment.

\section{\model}\label{sec:model}
\begin{figure}[htbp]
    \centering
    \includegraphics[width=1\linewidth]{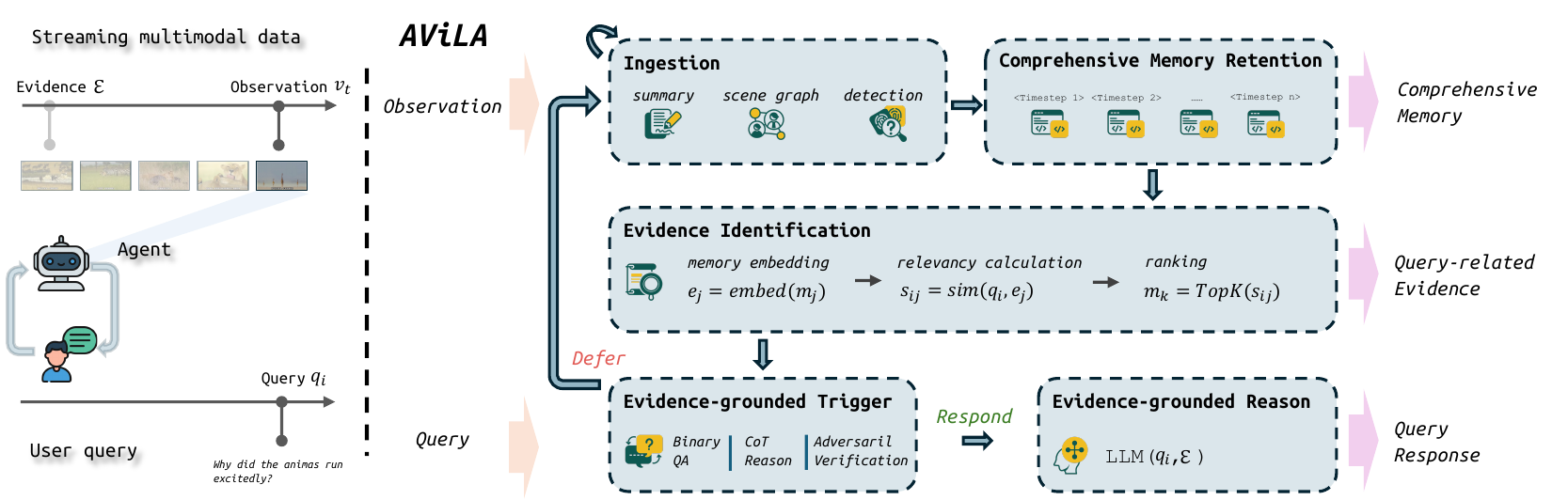}
    \caption{To tackle the tripartite challenges of increasing streaming data, ad-hoc user queries, and time-aware responses, we highlight three components of \model: (1) \textbf{Comprehensive Memory Retention}: ingested inputs are retained as a temporally indexed memory buffer, enabling access to historical observations for future queries; (2) \textbf{Evidence Identification}: Upon receiving a user query, the agent searches its memory for supporting evidence; (3) \textbf{Evidence-Grounded Trigger}: Based on the identified evidence, the agent decides whether to respond immediately or defer the answer.
}
    \label{fig:framework}
\end{figure}

We present \textbf{AViLA}, a framework designed to handle \qea when interacting with streaming multimodal data. 
AViLA operates in a temporally evolving setting where user queries may arrive at any time, before or after the relevant evidence is observed, and responds to the query asynchronously. 
As shown in Fig.~\ref{fig:framework}, \model is built around five core components, which are detailed in the following sections.

\subsection{Agent Pipeline}
\model begins with an \textbf{Ingestion} step, where frame-wise visual inputs are parsed into structured representations. This stage is designed to be \emph{query-agnostic}, since future queries are unknown at the time of observation. Following established practices~\citep{wang2025videoagent,wang2024videollamblongcontextvideounderstanding,zhang2024mm}, we apply a chain of perception tools, such as image captioning, scene graph extraction, and object detection, to extract a structured and compact representation from each frame. 
Importantly, this step must balance informativeness and efficiency: coarse representations may result in insufficient memory for future queries, while overly detailed ones incur significant computational overhead. 

\vspace{0.2cm}
\textbf{\memicon Comprehensive Memory Retention}
In the streaming setting, future questions are unpredictable and may reference past or ongoing events.
Thus, the agent must maintain a comprehensive and question-agnostic memory $M_t$ that effectively captures and organizes salient information from the video stream without prior knowledge of which details will become relevant.

We consider three types of memory formation: \textit{vision} memory, \textit{text} memory, and \textit{vision-text} memory.
For vision memory, we cache the raw frame in the memory bank to maintain fidelity. For text memory, we employ different tools for the ingestion step and save the textual descriptions of the past scenes.
Additionally, we consider \textit{object}-level memory extracted by MLLMs to accomplish more queries demanding object-level understanding.

\vspace{0.2cm}
\textbf{\evidenceicon Evidence Identification}
Maintaining an exhaustive and increasing memory of all past observations is computationally impractical for MLLMs and often unnecessary for effective reasoning. Instead, we explore different strategies for selectively identifying relevant evidence from the agent’s memory to support an arbitrary query. Specifically, we consider the following approaches:



\textbf{\textit{Relevancy-based identification}} The agent identifies evidence by retrieving memory entries that are most semantically aligned with the current query. 
For each memory snapshot $m_j \in \mathcal{M}$, we save the snapshot embeddings in the vector database.
For each query $q_i$, we compute a similarity score $s_j = \mathrm{sim}(q_i, m_j)$ between the query and each memory snapshot $m_j \in \mathcal{M}_t$. The top-K most relevant entries are selected to form the evidence set:
$\mathcal{E}{i} = \mathrm{TopK}_j(s_j)$.
This approach allows the agent to focus on the most contextually relevant past observations, even if they are temporally distant, enabling flexible reasoning across time. Retrieval can be implemented using either fixed similarity metrics (\eg, cosine similarity in the embedding space) or learned retrievers.

\vspace{0.2cm}
\textbf{\triggericon Evidence-Grounded Trigger}
To enable MLLMs to decide \emph{when} to answer a question in streaming video, we introduce a set of trigger strategies that assess the model’s \emph{readiness} given current memory and observation. A well-designed trigger adheres to two core principles: (1) It must respond \emph{promptly} when sufficient evidence is available, and (2) it must defer or refuse to answer when evidence is incomplete or absent. Guided by these principles, we implement and compare four distinct triggering methods for handling \qea.
Triggers are invoked upon the arrival of every new observation.

\textbf{\textit{Binary QA}} 
We formulate the trigger decision as a binary question-answering task. The model is prompted with the original query and current evidence context, and asked a yes/no question such as “\texttt{Is there sufficient information to answer this query?}”. The model’s binary response (\texttt{yes}/\texttt{no}) is used as the trigger signal to either proceed with answering or defer.

\textbf{\textit{CoT Evidence Reasoning}} 
To promote more deliberate evaluation of evidence sufficiency, we elicit a Chain-of-Thought (CoT)~\cite{chen2024measuring, wei2022chain} reasoning from the agent. The agent is prompted to reason explicitly over the available evidence $\mathcal{E}$, step by step, before concluding whether it is ready to answer. This encourages the model to reflect on what information is observed and what may still be missing.

\textbf{\textit{Adversarial Verification}} 
To balance between premature answering and excessive refusal, we propose a two-step adversarial protocol. In the first step, the model is asked whether the question is answerable under the current memory and perception context. In the second step, it is asked whether the query should be rejected. The trigger decision is based on consistency between these two outputs: If both agree, the decision is taken accordingly; if they disagree, the model defers to avoid uncertain or unsupported answers.

It is important to note that we do \emph{not} provide candidate answer options to the model when implementing the trigger. This avoids oversimplifying the task and prevents the model from exploiting shortcuts based on answer cues. Additional details are provided in Appendix Sec.~\ref{sec:more-model}.
Once the trigger determines that sufficient evidence has been gathered, \model proceeds to generate an answer grounded in that evidence. We provide the implementation details in Appendix Sec.~\ref{sec:more-ex}.

\subsection{Implementation Details}
We evaluate our framework using four Multimodal Large Language Models (MLLMs): \textit{LLaVA-NeXT-Video}~\cite{zhang2024llavanextvideo}, \textit{LLaVA-OneVision}~\cite{li2024llavaonevisioneasyvisualtask}, \textit{Qwen2.5-VL}~\cite{bai2025qwen2}, and \textit{Qwen2-VL-Instruct}~\cite{wang2024qwen2}. To simulate real-time processing, all models operate on video streams sampled at approximately 1 FPS. We buffer every 32 consecutive frames as a chunked observation and feed them densely into the model, ensuring full coverage of each 30-second video clip. All experiments are conducted using 2 NVIDIA A100 40GB GPUs, with model inference served through vLLM~\cite{kwon2023efficient} for efficient throughput. For object memory, we employ an off-the-shelf object detector~\cite{shen2024aligning} to extract bounding boxes and metadata, which are then converted into structured, rule-based scene graphs. For textual memory, we use the same MLLM backbone to caption 32-frame segments during the ingestion phase. All memory entries are stored in a FAISS~\cite{johnson2017billion} vector database to enable efficient similarity-based retrieval. Additional implementation details are provided in the Appendix.

\section{Experimental Study}\label{sec:ex}

\begin{wrapfigure}{r}{0.5\linewidth}
    \centering
    \includegraphics[width=\linewidth]{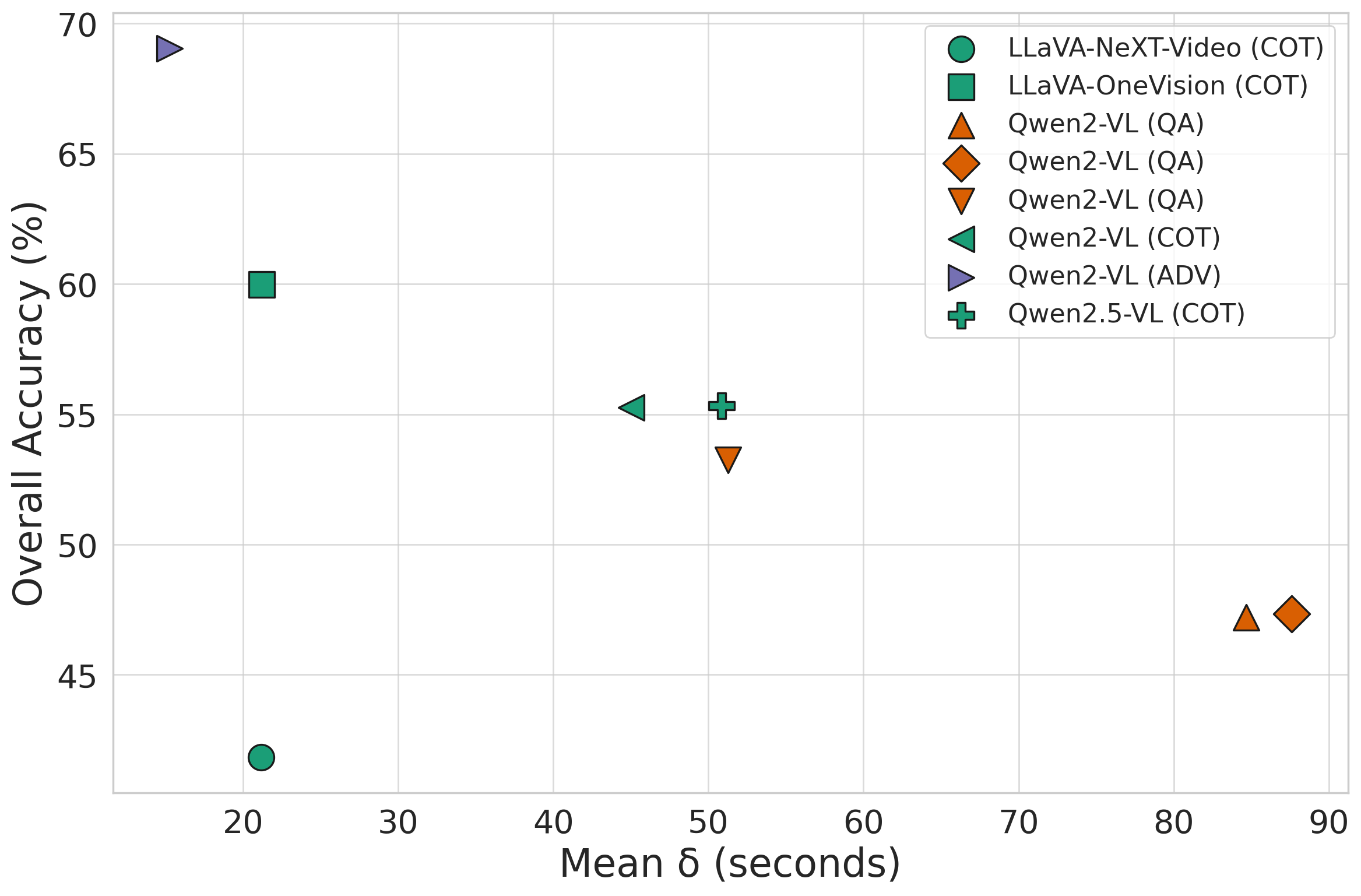}
    \caption{We plot overall accuracy versus mean temporal offsets across models and trigger types. Higher accuracy with shorter delay indicates better temporal awareness.}
    \label{fig:scatter}
\end{wrapfigure}

\vspace{0.2cm}
\textbf{Main Results}
We illustrate their performance in Fig.~\ref{fig:scatter}, comparing baseline MLLMs adapted to \dataset, and report the overall performance and more ablation studies in the Appendix (Sec.~\ref {sec:more-model}). We will elucidate the results in each dimension.

Among all configurations, the adversarial trigger consistently yields superior performance, achieving the highest accuracy (69.05\%) with the lowest mean offset (15.26s), indicating its effectiveness in balancing early answering with evidence sufficiency. CoT-based triggers demonstrate improved alignment over basic QA triggers, reducing hallucinations by encouraging deliberate reasoning. However, they incur moderate temporal offsets (mean $\delta \approx 45s$). In contrast, QA-triggered responses, while faster to activate, often result in delayed answers with high offset values exceeding 80s, suggesting inadequate temporal awareness. These trends hold consistently across MLLMs, with Qwen2-VL and LLaVA-OneVision exhibiting strong capabilities when coupled with CoT-based triggers. The results validate our agent design: ingesting dense multimodal memory, retrieving evidence via semantic similarity, and deferring responses until relevant evidence accumulates significantly enhances both temporal and factual grounding in streaming scenarios.



\vspace{0.2cm}
\textbf{Impact of Memory Modality on QA Accuracy and Timing}
Tab.~\ref{tab:memory} presents an ablation study on memory formation strategies for QA-triggered responses using Qwen2-VL-7B-Instruct over 30-second video clips. We evaluate four configurations: single-modal memory (vision, text) and two multimodal variants (text+vision, text+object).

Single-modality memory exhibits limited performance, with vision outperforming text in present and hypothetical queries, while text shows marginal gains on \futurefactq ones. However, both setups struggle with temporal awareness, yielding high temporal offset of responses (mean $\delta > 70$s).

Multimodal memory significantly improves performance. The text+vision configuration enhances overall accuracy (39.6\%) but introduces high temporal variance (std $\delta = 258.1$s), likely due to redundancy and semantic overlap. In contrast, the text+object memory achieves the highest accuracy (41.3\%) and the lowest temporal offset (mean $\delta = 36.6$s), demonstrating its ability to support precise, time-aware reasoning. This highlight the importance of structured memory for model responses with temporally dispersed evidence.

\begin{table}[!h]
\centering
\scriptsize
\setlength{\tabcolsep}{3pt}
\caption{Performance of Qwen2-VL-7B with QA trigger across different memory configurations.}
\begin{tabular}{l|ccccc|cc}
\toprule
\multirow{2}{*}{\textbf{Memory}} & \multicolumn{5}{c|}{\textbf{Accuracy (\%)}} & \multicolumn{2}{c}{\textbf{Temporal Offset (s)}} \\
 & \textbf{Overall}↑ & \pastq↑ & \presentq↑ & \futurehypoq↑ & \futurefactq↑ & \textbf{Mean $\delta$}$\downarrow$ & \textbf{Std $\delta$}$\downarrow$ \\
\midrule
vision         & 39.2 & \textbf{37.6} & 43.0 & 43.9 & 27.1 & 70.7 & 179.8 \\
text           & 33.8 & 29.2 & 33.7 & 41.7 & 34.1 & 84.7 & 253.1 \\
text+vision    & 39.6 & 38.3 & 42.5 & 37.1 & 38.2 & 87.6 & 258.1 \\
text+object    & \textbf{41.3} & 28.4 & \textbf{49.1} & \textbf{46.3} & \textbf{45.1} & \textbf{36.6} & \textbf{95.5} \\
\bottomrule
\end{tabular}
\label{tab:memory}
\end{table}

\textbf{Effectiveness of Trigger Strategies}  
Tab.~\ref{tab:trigger} evaluates the impact of different evidence-grounded triggering mechanisms on QA performance using text memory. The \textit{Binary QA} trigger achieves modest accuracy (33.8\%) with relatively high response latency (mean $\delta = 84.7$s), indicating that simple binary prompting may be insufficient for effective temporal awareness. In contrast, the \textit{CoT Evidence Reasoning} trigger significantly improves accuracy (55.3\%) while reducing the temporal offset of responses (mean $\delta = 45.0$s), suggesting that intermediate reasoning helps the model assess evidence sufficiency more effectively. The \textit{Adversarial Verification} trigger yields the strongest overall performance, achieving the highest accuracy (61.5\%) and consistently outperforming other methods across all temporal categories. It also produces the most temporally aligned responses, with the lowest mean temporal offset ($17.3$s) and variance ($20.6$s). These results underscore the benefits of verification-based prompting for improving both response quality and timing in streaming QA.

\begin{table}[!h]
\centering
\scriptsize
\setlength{\tabcolsep}{3pt} 
\caption{
Performance using different evidence-grounded triggers for Qwen2-VL-7B.}
\begin{tabular}{l|ccccc|cc}
\toprule
\multirow{2}{*}{\textbf{Trigger}} & \multicolumn{5}{c|}{\textbf{Accuracy (\%)}} & \multicolumn{2}{c}{\textbf{Temporal Offset (s)}} \\
 & \textbf{Overall}↑ & \pastq↑ & \presentq↑ & \futurehypoq↑ & \futurefactq↑ & \textbf{Mean $\delta$}$\downarrow$ & \textbf{Std $\delta$}$\downarrow$ \\
\midrule
Binary QA                  & 33.8 & 29.2 & 33.7 & 41.7 & 34.1 & 84.7  & 253.1 \\
CoT Evidence Reasoning     & 55.3 & 43.8 & 61.7 & 54.1 & \textbf{60.9} & 45.0  & 115.9 \\
Adversarial Verification   & \textbf{61.5} & \textbf{45.6} & \textbf{73.8} & \textbf{70.1} & 55.2 & \textbf{17.3} & \textbf{20.6} \\
\bottomrule
\end{tabular}
\label{tab:trigger}
\end{table}

\vspace{0.2em}
\textbf{Performance Across Temporal Categories}  
We analyze how models handle questions whose supporting evidence appears in the \textit{past}, \textit{present}, or \textit{future}, following our temporal categorization. Results in Tab.s~\ref{tab:memory}--\ref{tab:llm} reveal consistent trends across memory configurations, trigger strategies, and backbone MLLMs. \presentq queries consistently achieve the highest accuracy, benefiting from direct alignment with the visible clip. For instance, \textit{Adversarial Verification} reaches 73.8\% accuracy on these queries. Both \futurefactq and \futurehypoq are also handled effectively by setups with strong reasoning capabilities. \textit{Qwen2.5-VL-3B} reaches 72.0\% on \futurehypoq queries, while \textit{Adversarial Verification} attains 70.1\% and 55.2\% on \futurehypoq and \futurefactq queries, respectively. This suggests that reasoning-oriented triggers and pretrained LLM alignment help anticipate future stream and identify the lack of evidence. \pastq queries, however, remain the most challenging across all settings. Even the best-performing configurations, such as \textit{Adversarial Verification}, achieve only 45.6\%. Simpler baselines, such as text-only memory with QA triggers, perform poorly (\eg, 29.2\%), underscoring the difficulty of retaining and utilizing temporally distant evidence. Overall, these findings emphasize the critical role of memory fidelity, trigger design, and LLM reasoning alignment in handling temporally disjoint evidence in streaming QA tasks.

\vspace{0.2cm}
\textbf{Impact of Backbone MLLMs}  
Tab.~\ref{tab:llm} compares different backbone MLLMs under a consistent setup using CoT Evidence Reasoning triggers, text memory, and 30-second video clips. \textit{LLaVA-OneVision-7B} achieves the best overall accuracy (59.1\%) with strong performance across all temporal categories and minimal temporal offset of responses (mean $\delta = 21.2$s), indicating effective temporal grounding and evidence-based reasoning. \textit{Qwen2.5-VL-3B} performs competitively (55.3\%), particularly excelling in \futurehypoq queries (72.0\%), but exhibits higher latency (mean $\delta = 50.9$s), which suggests delayed readiness to answer. \textit{Qwen2-VL-7B} also achieves 55.3\% accuracy, with balanced performance and moderate temporal offset, while \textit{LLaVA-NeXT-Video-7B} underperforms in accuracy (41.4\%), especially on past and factual queries, despite maintaining a low average temporal offset (21.2s). These findings highlight that model scale alone does not guarantee strong performance; the alignment of pretraining objectives and reasoning capabilities critically influences both accuracy and temporal responsiveness in streaming QA tasks.

\begin{table}[!h]
\centering
\scriptsize
\setlength{\tabcolsep}{3pt}
\caption{Performance of different MLLMs with CoT Reasoning triggers and textual memory.}
\begin{tabular}{l|ccccc|cc}
\toprule
\multirow{2}{*}{\textbf{MLLM}} & \multicolumn{5}{c|}{\textbf{Accuracy (\%)}} & \multicolumn{2}{c}{\textbf{Temporal Offset (s)}} \\
 & \textbf{Overall}↑ & \pastq↑ & \presentq↑ & \futurehypoq↑ & \futurefactq↑ & \textbf{Mean $\delta$}$\downarrow$ & \textbf{Std $\delta$}$\downarrow$ \\
\midrule
Qwen2-VL-7B         & 55.3  & \textbf{43.8} & 61.7 & 54.1 & \textbf{60.9} & 45.0  & 115.9 \\
Qwen2.5-VL-3B       & 55.3  & 38.4 & \textbf{62.5} & \textbf{72.0} & 54.3 & 50.9  & 128.3 \\
LLaVA-NeXT-Video-7B & 41.4  & 30.4 & 48.2 & 45.5 & 40.6 & 21.2 & 72.3 \\
LLaVA-OneVision-7B  & \textbf{59.1} & \textbf{43.8} & 69.7 & 67.6 & 53.8 & \textbf{21.2} & \textbf{72.2} \\
\bottomrule
\end{tabular}
\label{tab:llm}
\end{table}

\section{Conclusion}
In this work, we introduce and study \qea, a key challenge that user queries and their supporting evidence appear asynchronously in streaming agent interaction. We present a diagnostic dataset, \dataset, to systematically explore how agents are confronted with \qea. To tackle ad-hoc user queries in streaming settings and elicit time-aware responses, we propose \model, an agentic framework to identify supporting evidence and process asynchronous queries. \model demonstrates non-marginal performance compared to vanilla baseline methods and presents a solid baseline for interaction with streaming multimodal data.

\textbf{Limitations}
Our framework relies on pretrained MLLMs and operates solely at inference time, without fine-tuning or preference optimization. While this enables general applicability, it may limit calibration with the measure of evidence sufficiency. Future work could explore more adaptive and explainable trigger mechanisms to improve response calibration and temporal awareness.

\textbf{Societal Impacts}
This work contributes to the development of time-aware multimodal agents, which may enhance real-world applications.
While the proposed framework improves reasoning and temporal awareness in streaming multimodal data, misuse of time-aware response systems in surveillance or behavioral prediction could amplify bias or erode privacy. 

{
\small
\bibliographystyle{plainnat}
\bibliography{references}

\begin{thebibliography}{79}
\providecommand{\natexlab}[1]{#1}
\providecommand{\url}[1]{\texttt{#1}}
\expandafter\ifx\csname urlstyle\endcsname\relax
  \providecommand{\doi}[1]{doi: #1}\else
  \providecommand{\doi}{doi: \begingroup \urlstyle{rm}\Url}\fi

\bibitem[Alayrac et~al.(2022)Alayrac, Donahue, Luc, Miech, Barr, Hasson, Lenc, Mensch, Millican, Reynolds, et~al.]{alayrac2022flamingo}
Jean-Baptiste Alayrac, Jeff Donahue, Pauline Luc, Antoine Miech, Iain Barr, Yana Hasson, Karel Lenc, Arthur Mensch, Katherine Millican, Malcolm Reynolds, et~al.
\newblock Flamingo: a visual language model for few-shot learning.
\newblock \emph{Advances in neural information processing systems}, 35:\penalty0 23716--23736, 2022.

\bibitem[Anthropic(2024)]{claude3report2024}
Anthropic.
\newblock Claude: The 2024 anthropic model family.
\newblock \url{https://www.anthropic.com/news/claude-3-family}, 2024.

\bibitem[Argaw et~al.(2024)Argaw, Yoon, Heilbron, Deilamsalehy, Bui, Wang, Dernoncourt, and Chung]{argaw2024scaling}
Dawit~Mureja Argaw, Seunghyun Yoon, Fabian~Caba Heilbron, Hanieh Deilamsalehy, Trung Bui, Zhaowen Wang, Franck Dernoncourt, and Joon~Son Chung.
\newblock Scaling up video summarization pretraining with large language models.
\newblock In \emph{Proceedings of the IEEE/CVF Conference on Computer Vision and Pattern Recognition}, pages 8332--8341, 2024.

\bibitem[Bai et~al.(2023)Bai, Bai, Yang, Wang, Tan, Wang, Lin, Zhou, and Zhou]{bai2023qwen}
Jinze Bai, Shuai Bai, Shusheng Yang, Shijie Wang, Sinan Tan, Peng Wang, Junyang Lin, Chang Zhou, and Jingren Zhou.
\newblock Qwen-vl: A frontier large vision-language model with versatile abilities.
\newblock \emph{arXiv preprint arXiv:2308.12966}, 1\penalty0 (2):\penalty0 3, 2023.

\bibitem[Bai et~al.(2025)Bai, Chen, Liu, Wang, Ge, Song, Dang, Wang, Wang, Tang, et~al.]{bai2025qwen2}
Shuai Bai, Keqin Chen, Xuejing Liu, Jialin Wang, Wenbin Ge, Sibo Song, Kai Dang, Peng Wang, Shijie Wang, Jun Tang, et~al.
\newblock Qwen2. 5-vl technical report.
\newblock \emph{arXiv preprint arXiv:2502.13923}, 2025.

\bibitem[Bain et~al.(2021)Bain, Nagrani, Varol, and Zisserman]{bain2021frozen}
Max Bain, Arsha Nagrani, G{\"u}l Varol, and Andrew Zisserman.
\newblock Frozen in time: A joint video and image encoder for end-to-end retrieval.
\newblock In \emph{Proceedings of the IEEE/CVF international conference on computer vision}, pages 1728--1738, 2021.

\bibitem[Barrios et~al.(2023)Barrios, Soldan, Ceballos-Arroyo, Heilbron, and Ghanem]{barrios2023localizing}
Wayner Barrios, Mattia Soldan, Alberto~Mario Ceballos-Arroyo, Fabian~Caba Heilbron, and Bernard Ghanem.
\newblock Localizing moments in long video via multimodal guidance.
\newblock In \emph{Proceedings of the IEEE/CVF International Conference on Computer Vision}, pages 13667--13678, 2023.

\bibitem[Bi et~al.(2024)Bi, Chen, Chen, Chen, Dai, Deng, Ding, Dong, Du, Fu, et~al.]{bi2024deepseek}
Xiao Bi, Deli Chen, Guanting Chen, Shanhuang Chen, Damai Dai, Chengqi Deng, Honghui Ding, Kai Dong, Qiushi Du, Zhe Fu, et~al.
\newblock Deepseek llm: Scaling open-source language models with longtermism.
\newblock \emph{arXiv preprint arXiv:2401.02954}, 2024.

\bibitem[Chen et~al.(2024{\natexlab{a}})Chen, Lv, Wu, Lin, Song, Gao, Liu, Gao, Mao, and Shou]{Chen_2024_CVPR}
Joya Chen, Zhaoyang Lv, Shiwei Wu, Kevin~Qinghong Lin, Chenan Song, Difei Gao, Jia-Wei Liu, Ziteng Gao, Dongxing Mao, and Mike~Zheng Shou.
\newblock Videollm-online: Online video large language model for streaming video.
\newblock In \emph{Proceedings of the IEEE/CVF Conference on Computer Vision and Pattern Recognition (CVPR)}, pages 18407--18418, June 2024{\natexlab{a}}.

\bibitem[Chen et~al.(2023)Chen, Zhu, Shen, Li, Liu, Zhang, Krishnamoorthi, Chandra, Xiong, and Elhoseiny]{chen2023minigpt}
Jun Chen, Deyao Zhu, Xiaoqian Shen, Xiang Li, Zechun Liu, Pengchuan Zhang, Raghuraman Krishnamoorthi, Vikas Chandra, Yunyang Xiong, and Mohamed Elhoseiny.
\newblock Minigpt-v2: large language model as a unified interface for vision-language multi-task learning.
\newblock \emph{arXiv preprint arXiv:2310.09478}, 2023.

\bibitem[Chen et~al.(2024{\natexlab{b}})Chen, Wei, Li, Dong, Zhang, Zang, Chen, Duan, Tang, Yuan, et~al.]{chen2024sharegpt4video}
Lin Chen, Xilin Wei, Jinsong Li, Xiaoyi Dong, Pan Zhang, Yuhang Zang, Zehui Chen, Haodong Duan, Zhenyu Tang, Li~Yuan, et~al.
\newblock Sharegpt4video: Improving video understanding and generation with better captions.
\newblock \emph{Advances in Neural Information Processing Systems}, 37:\penalty0 19472--19495, 2024{\natexlab{b}}.

\bibitem[Chen et~al.(2024{\natexlab{c}})Chen, Sikka, Cogswell, Ji, and Divakaran]{chen2024measuring}
Yangyi Chen, Karan Sikka, Michael Cogswell, Heng Ji, and Ajay Divakaran.
\newblock Measuring and improving chain-of-thought reasoning in vision-language models.
\newblock In \emph{Proceedings of the 2024 Conference of the North American Chapter of the Association for Computational Linguistics: Human Language Technologies (Volume 1: Long Papers)}, pages 192--210, 2024{\natexlab{c}}.

\bibitem[Chen et~al.(2024{\natexlab{d}})Chen, Li, Bao, Patel, Kong, Min, and Metaxas]{chen2024learning}
Yuxiao Chen, Kai Li, Wentao Bao, Deep Patel, Yu~Kong, Martin~Renqiang Min, and Dimitris~N Metaxas.
\newblock Learning to localize actions in instructional videos with llm-based multi-pathway text-video alignment.
\newblock In \emph{European Conference on Computer Vision}, pages 193--210. Springer, 2024{\natexlab{d}}.

\bibitem[Chen et~al.(2024{\natexlab{e}})Chen, Wu, Wang, Su, Chen, Xing, Zhong, Zhang, Zhu, Lu, et~al.]{chen2024internvl}
Zhe Chen, Jiannan Wu, Wenhai Wang, Weijie Su, Guo Chen, Sen Xing, Muyan Zhong, Qinglong Zhang, Xizhou Zhu, Lewei Lu, et~al.
\newblock Internvl: Scaling up vision foundation models and aligning for generic visual-linguistic tasks.
\newblock In \emph{Proceedings of the IEEE/CVF conference on computer vision and pattern recognition}, pages 24185--24198, 2024{\natexlab{e}}.

\bibitem[Cheng et~al.(2024)Cheng, Leng, Zhang, Xin, Li, Chen, Zhu, Zhang, Luo, Zhao, et~al.]{cheng2024videollama}
Zesen Cheng, Sicong Leng, Hang Zhang, Yifei Xin, Xin Li, Guanzheng Chen, Yongxin Zhu, Wenqi Zhang, Ziyang Luo, Deli Zhao, et~al.
\newblock Videollama 2: Advancing spatial-temporal modeling and audio understanding in video-llms.
\newblock \emph{arXiv preprint arXiv:2406.07476}, 2024.

\bibitem[Driess et~al.(2023)Driess, Xia, Sajjadi, Lynch, Chowdhery, Wahid, Tompson, Vuong, Yu, Huang, et~al.]{driess2023palm}
Danny Driess, Fei Xia, Mehdi~SM Sajjadi, Corey Lynch, Aakanksha Chowdhery, Ayzaan Wahid, Jonathan Tompson, Quan Vuong, Tianhe Yu, Wenlong Huang, et~al.
\newblock Palm-e: An embodied multimodal language model.
\newblock 2023.

\bibitem[Fan et~al.(2024{\natexlab{a}})Fan, Ma, Wu, Du, Li, Gao, and Li]{fan2024videoagent}
Yue Fan, Xiaojian Ma, Rujie Wu, Yuntao Du, Jiaqi Li, Zhi Gao, and Qing Li.
\newblock Videoagent: A memory-augmented multimodal agent for video understanding.
\newblock In \emph{European Conference on Computer Vision}, pages 75--92. Springer, 2024{\natexlab{a}}.

\bibitem[Fan et~al.(2024{\natexlab{b}})Fan, Ma, Wu, Du, Li, Gao, and Li]{fan2024videoagentmemoryaugmentedmultimodalagent}
Yue Fan, Xiaojian Ma, Rujie Wu, Yuntao Du, Jiaqi Li, Zhi Gao, and Qing Li.
\newblock Videoagent: A memory-augmented multimodal agent for video understanding, 2024{\natexlab{b}}.
\newblock URL \url{https://arxiv.org/abs/2403.11481}.

\bibitem[Fu et~al.(2024)Fu, Dai, Luo, Li, Ren, Zhang, Wang, Zhou, Shen, Zhang, Chen, Li, Lin, Zhao, Li, Xu, Zheng, Chen, Ji, and Sun]{Video-MME}
Chaoyou Fu, Yuhan Dai, Yongdong Luo, Lei Li, Shuhuai Ren, Renrui Zhang, Zihan Wang, Chenyu Zhou, Yunhang Shen, Mengdan Zhang, Peixian Chen, Yanwei Li, Shaohui Lin, Sirui Zhao, Ke~Li, Tong Xu, Xiawu Zheng, Enhong Chen, Rongrong Ji, and Xing Sun.
\newblock Video-mme: The first-ever comprehensive evaluation benchmark of multi-modal llms in video analysis, 2024.
\newblock URL \url{https://arxiv.org/abs/2405.21075}.

\bibitem[Gao et~al.(2023)Gao, Zhou, Ji, Zhu, Yang, and Shou]{gao2023mist}
Difei Gao, Luowei Zhou, Lei Ji, Linchao Zhu, Yi~Yang, and Mike~Zheng Shou.
\newblock Mist: Multi-modal iterative spatial-temporal transformer for long-form video question answering.
\newblock In \emph{Proceedings of the IEEE/CVF conference on computer vision and pattern recognition}, pages 14773--14783, 2023.

\bibitem[Grauman et~al.(2022)Grauman, Westbury, Byrne, Chavis, Furnari, Girdhar, Hamburger, Jiang, Liu, Liu, Martin, Nagarajan, Radosavovic, Ramakrishnan, Ryan, Sharma, Wray, Xu, Xu, Zhao, Bansal, Batra, Cartillier, Crane, Do, Doulaty, Erapalli, Feichtenhofer, Fragomeni, Fu, Gebreselasie, González, Hillis, Huang, Huang, Jia, Khoo, Koláĭ, Kottur, Kumar, Landini, Li, Li, Li, Mangalam, Modhugu, Munro, Murrell, Nishiyasu, Price, Puentes, Ramazanova, Sari, Somasundaram, Southerland, Sugano, Tao, Vo, Wang, Wu, Yagi, Zhao, Zhu, Arbeláez, Crandall, Damen, Farinella, Fuegen, Ghanem, Ithapu, Jawahar, Joo, Kitani, Li, Newcombe, Oliva, Park, Rehg, Sato, Shi, Shou, Torralba, Torresani, Yan, and Malik]{Ego4D}
Kristen Grauman, Andrew Westbury, Eugene Byrne, Zachary Chavis, Antonino Furnari, Rohit Girdhar, Jackson Hamburger, Hao Jiang, Miao Liu, Xingyu Liu, Miguel Martin, Tushar Nagarajan, Ilija Radosavovic, Santhosh~Kumar Ramakrishnan, Fiona Ryan, Jayant Sharma, Michael Wray, Mengmeng Xu, Eric~Zhongcong Xu, Chen Zhao, Siddhant Bansal, Dhruv Batra, Vincent Cartillier, Sean Crane, Tien Do, Morrie Doulaty, Akshay Erapalli, Christoph Feichtenhofer, Adriano Fragomeni, Qichen Fu, Abrham Gebreselasie, Cristina González, James Hillis, Xuhua Huang, Yifei Huang, Wenqi Jia, Weslie Khoo, Jáchym Koláĭ, Satwik Kottur, Anurag Kumar, Federico Landini, Chao Li, Yanghao Li, Zhenqiang Li, Karttikeya Mangalam, Raghava Modhugu, Jonathan Munro, Tullie Murrell, Takumi Nishiyasu, Will Price, Paola~Ruiz Puentes, Merey Ramazanova, Leda Sari, Kiran Somasundaram, Audrey Southerland, Yusuke Sugano, Ruijie Tao, Minh Vo, Yuchen Wang, Xindi Wu, Takuma Yagi, Ziwei Zhao, Yunyi Zhu, Pablo Arbeláez, David Crandall, Dima Damen, Giovanni~Maria
  Farinella, Christian Fuegen, Bernard Ghanem, Vamsi~Krishna Ithapu, C.~V. Jawahar, Hanbyul Joo, Kris Kitani, Haizhou Li, Richard Newcombe, Aude Oliva, Hyun~Soo Park, James~M. Rehg, Yoichi Sato, Jianbo Shi, Mike~Zheng Shou, Antonio Torralba, Lorenzo Torresani, Mingfei Yan, and Jitendra Malik.
\newblock Ego4d: Around the world in 3,000 hours of egocentric video.
\newblock In \emph{2022 IEEE/CVF Conference on Computer Vision and Pattern Recognition (CVPR)}, pages 18973--18990, 2022.
\newblock \doi{10.1109/CVPR52688.2022.01842}.

\bibitem[Guo et~al.(2025)Guo, Liu, Li, Cheng, Tang, Sui, Liu, Chen, and Zhao]{guo2025vtg}
Yongxin Guo, Jingyu Liu, Mingda Li, Dingxin Cheng, Xiaoying Tang, Dianbo Sui, Qingbin Liu, Xi~Chen, and Kevin Zhao.
\newblock Vtg-llm: Integrating timestamp knowledge into video llms for enhanced video temporal grounding.
\newblock In \emph{Proceedings of the AAAI Conference on Artificial Intelligence}, volume~39, pages 3302--3310, 2025.

\bibitem[Hong et~al.(2024)Hong, Wang, Lv, Xu, Yu, Ji, Wang, Wang, Dong, Ding, et~al.]{hong2024cogagent}
Wenyi Hong, Weihan Wang, Qingsong Lv, Jiazheng Xu, Wenmeng Yu, Junhui Ji, Yan Wang, Zihan Wang, Yuxiao Dong, Ming Ding, et~al.
\newblock Cogagent: A visual language model for gui agents.
\newblock In \emph{Proceedings of the IEEE/CVF Conference on Computer Vision and Pattern Recognition}, pages 14281--14290, 2024.

\bibitem[Hu et~al.(2024)Hu, Tu, Han, He, Cui, Long, Zheng, Fang, Huang, Zhao, et~al.]{hu2024minicpm}
Shengding Hu, Yuge Tu, Xu~Han, Chaoqun He, Ganqu Cui, Xiang Long, Zhi Zheng, Yewei Fang, Yuxiang Huang, Weilin Zhao, et~al.
\newblock Minicpm: Unveiling the potential of small language models with scalable training strategies.
\newblock \emph{arXiv preprint arXiv:2404.06395}, 2024.

\bibitem[Hua et~al.(2025)Hua, Tang, Xu, and Luo]{hua2025v2xum}
Hang Hua, Yunlong Tang, Chenliang Xu, and Jiebo Luo.
\newblock V2xum-llm: Cross-modal video summarization with temporal prompt instruction tuning.
\newblock In \emph{Proceedings of the AAAI Conference on Artificial Intelligence}, volume~39, pages 3599--3607, 2025.

\bibitem[Hurst et~al.(2024)Hurst, Lerer, Goucher, Perelman, Ramesh, Clark, Ostrow, Welihinda, Hayes, Radford, et~al.]{hurst2024gpt}
Aaron Hurst, Adam Lerer, Adam~P Goucher, Adam Perelman, Aditya Ramesh, Aidan Clark, AJ~Ostrow, Akila Welihinda, Alan Hayes, Alec Radford, et~al.
\newblock Gpt-4o system card.
\newblock \emph{arXiv preprint arXiv:2410.21276}, 2024.

\bibitem[Johnson et~al.(2017)Johnson, Douze, and J{\'e}gou]{johnson2017billion}
Jeff Johnson, Matthijs Douze, and Herv{\'e} J{\'e}gou.
\newblock Billion-scale similarity search with gpus. corr abs/1702.08734 (2017).
\newblock \emph{arXiv preprint arXiv:1702.08734}, 2017.

\bibitem[Kahatapitiya et~al.(2025)Kahatapitiya, Ranasinghe, Park, and Ryoo]{kahatapitiya2025language}
Kumara Kahatapitiya, Kanchana Ranasinghe, Jongwoo Park, and Michael~S Ryoo.
\newblock Language repository for long video understanding.
\newblock In \emph{Workshop on Video-Language Models @ NeurIPS 2024}, 2025.
\newblock URL \url{https://openreview.net/forum?id=CTaWnvcTNz}.

\bibitem[Kwon et~al.(2023)Kwon, Li, Zhuang, Sheng, Zheng, Yu, Gonzalez, Zhang, and Stoica]{kwon2023efficient}
Woosuk Kwon, Zhuohan Li, Siyuan Zhuang, Ying Sheng, Lianmin Zheng, Cody~Hao Yu, Joseph~E. Gonzalez, Hao Zhang, and Ion Stoica.
\newblock Efficient memory management for large language model serving with pagedattention.
\newblock In \emph{Proceedings of the ACM SIGOPS 29th Symposium on Operating Systems Principles}, 2023.

\bibitem[Lei et~al.(2021)Lei, Li, Zhou, Gan, Berg, Bansal, and Liu]{lei2021less}
Jie Lei, Linjie Li, Luowei Zhou, Zhe Gan, Tamara~L Berg, Mohit Bansal, and Jingjing Liu.
\newblock Less is more: Clipbert for video-and-language learning via sparse sampling.
\newblock In \emph{Proceedings of the IEEE/CVF conference on computer vision and pattern recognition}, pages 7331--7341, 2021.

\bibitem[Li et~al.(2024{\natexlab{a}})Li, Zhang, Guo, Zhang, Li, Zhang, Zhang, Li, Liu, and Li]{li2024llavaonevisioneasyvisualtask}
Bo~Li, Yuanhan Zhang, Dong Guo, Renrui Zhang, Feng Li, Hao Zhang, Kaichen Zhang, Yanwei Li, Ziwei Liu, and Chunyuan Li.
\newblock Llava-onevision: Easy visual task transfer, 2024{\natexlab{a}}.
\newblock URL \url{https://arxiv.org/abs/2408.03326}.

\bibitem[Li et~al.(2024{\natexlab{b}})Li, Klabjan, and Utke]{li2024unsupervised}
Hanqing Li, Diego Klabjan, and Jean Utke.
\newblock Unsupervised video summarization via iterative training and simplified gan.
\newblock In \emph{Proceedings of the Asian Conference on Computer Vision}, pages 1585--1601, 2024{\natexlab{b}}.

\bibitem[Li et~al.(2023{\natexlab{a}})Li, Li, Savarese, and Hoi]{li2023blip}
Junnan Li, Dongxu Li, Silvio Savarese, and Steven Hoi.
\newblock Blip-2: Bootstrapping language-image pre-training with frozen image encoders and large language models.
\newblock In \emph{International conference on machine learning}, pages 19730--19742. PMLR, 2023{\natexlab{a}}.

\bibitem[Li et~al.(2023{\natexlab{b}})Li, He, Wang, Li, Wang, Luo, Wang, Wang, and Qiao]{li2023videochat}
KunChang Li, Yinan He, Yi~Wang, Yizhuo Li, Wenhai Wang, Ping Luo, Yali Wang, Limin Wang, and Yu~Qiao.
\newblock Videochat: Chat-centric video understanding.
\newblock \emph{arXiv preprint arXiv:2305.06355}, 2023{\natexlab{b}}.

\bibitem[Li et~al.(2025)Li, Niu, Miao, Ge, Zhou, He, Dong, Duan, Ding, Qian, Zhang, Zang, Cao, He, and Wang]{li2025ovobenchfarvideollmsrealworld}
Yifei Li, Junbo Niu, Ziyang Miao, Chunjiang Ge, Yuanhang Zhou, Qihao He, Xiaoyi Dong, Haodong Duan, Shuangrui Ding, Rui Qian, Pan Zhang, Yuhang Zang, Yuhang Cao, Conghui He, and Jiaqi Wang.
\newblock Ovo-bench: How far is your video-llms from real-world online video understanding?, 2025.
\newblock URL \url{https://arxiv.org/abs/2501.05510}.

\bibitem[Liao et~al.(2024)Liao, Li, Niu, and Zhang]{liao2024align}
Zhaohe Liao, Jiangtong Li, Li~Niu, and Liqing Zhang.
\newblock Align and aggregate: Compositional reasoning with video alignment and answer aggregation for video question-answering.
\newblock In \emph{Proceedings of the IEEE/CVF Conference on Computer Vision and Pattern Recognition}, pages 13395--13404, 2024.

\bibitem[Lin et~al.(2024)Lin, Fang, Chen, Wan, Luo, Li, Liu, and Sun]{lin2024streamingbenchassessinggapmllms}
Junming Lin, Zheng Fang, Chi Chen, Zihao Wan, Fuwen Luo, Peng Li, Yang Liu, and Maosong Sun.
\newblock Streamingbench: Assessing the gap for mllms to achieve streaming video understanding, 2024.
\newblock URL \url{https://arxiv.org/abs/2411.03628}.

\bibitem[Lin et~al.(2023)Lin, Ahmed, Li, Lin, Azarnasab, Yang, Wang, Liang, Liu, Lu, Liu, and Wang]{lin2023mmvidadvancingvideounderstanding}
Kevin Lin, Faisal Ahmed, Linjie Li, Chung-Ching Lin, Ehsan Azarnasab, Zhengyuan Yang, Jianfeng Wang, Lin Liang, Zicheng Liu, Yumao Lu, Ce~Liu, and Lijuan Wang.
\newblock Mm-vid: Advancing video understanding with gpt-4v(ision), 2023.
\newblock URL \url{https://arxiv.org/abs/2310.19773}.

\bibitem[Liu et~al.(2023{\natexlab{a}})Liu, Li, Wu, and Lee]{liu2023visual}
Haotian Liu, Chunyuan Li, Qingyang Wu, and Yong~Jae Lee.
\newblock Visual instruction tuning.
\newblock \emph{Advances in neural information processing systems}, 36:\penalty0 34892--34916, 2023{\natexlab{a}}.

\bibitem[Liu et~al.(2023{\natexlab{b}})Liu, Zhang, Liu, Dai, Yang, Ji, Feng, and Gong]{liu2023video}
Meng Liu, Mingda Zhang, Jialu Liu, Hanjun Dai, Ming-Hsuan Yang, Shuiwang Ji, Zheyun Feng, and Boqing Gong.
\newblock Video timeline modeling for news story understanding.
\newblock \emph{Advances in Neural Information Processing Systems}, 36:\penalty0 28294--28310, 2023{\natexlab{b}}.

\bibitem[Maaz et~al.(2023)Maaz, Rasheed, Khan, and Khan]{maaz2023video}
Muhammad Maaz, Hanoona Rasheed, Salman Khan, and Fahad~Shahbaz Khan.
\newblock Video-chatgpt: Towards detailed video understanding via large vision and language models.
\newblock \emph{arXiv preprint arXiv:2306.05424}, 2023.

\bibitem[Min et~al.(2024)Min, Buch, Nagrani, Cho, and Schmid]{min2024morevqa}
Juhong Min, Shyamal Buch, Arsha Nagrani, Minsu Cho, and Cordelia Schmid.
\newblock Morevqa: Exploring modular reasoning models for video question answering.
\newblock In \emph{Proceedings of the IEEE/CVF Conference on Computer Vision and Pattern Recognition}, pages 13235--13245, 2024.

\bibitem[Mu et~al.(2023)Mu, Zhang, Hu, Wang, Ding, Jin, Wang, Dai, Qiao, and Luo]{mu2023embodiedgpt}
Yao Mu, Qinglong Zhang, Mengkang Hu, Wenhai Wang, Mingyu Ding, Jun Jin, Bin Wang, Jifeng Dai, Yu~Qiao, and Ping Luo.
\newblock Embodiedgpt: Vision-language pre-training via embodied chain of thought.
\newblock \emph{Advances in Neural Information Processing Systems}, 36:\penalty0 25081--25094, 2023.

\bibitem[Nagrani et~al.(2024)Nagrani, Zhang, Mehran, Hornung, Gundavarapu, Jha, Myers, Zhou, Gong, Schmid, et~al.]{nagrani2024neptune}
Arsha Nagrani, Mingda Zhang, Ramin Mehran, Rachel Hornung, Nitesh~Bharadwaj Gundavarapu, Nilpa Jha, Austin Myers, Xingyi Zhou, Boqing Gong, Cordelia Schmid, et~al.
\newblock Neptune: The long orbit to benchmarking long video understanding.
\newblock \emph{arXiv preprint arXiv:2412.09582}, 2024.

\bibitem[Qian et~al.(2024)Qian, Dong, Zhang, Zang, Ding, Lin, and Wang]{qian2024streaming}
Rui Qian, Xiaoyi Dong, Pan Zhang, Yuhang Zang, Shuangrui Ding, Dahua Lin, and Jiaqi Wang.
\newblock Streaming long video understanding with large language models.
\newblock \emph{Advances in Neural Information Processing Systems}, 37:\penalty0 119336--119360, 2024.

\bibitem[Qian et~al.(2025)Qian, Ding, Dong, Zhang, Zang, Cao, Lin, and Wang]{qian2025dispider}
Rui Qian, Shuangrui Ding, Xiaoyi Dong, Pan Zhang, Yuhang Zang, Yuhang Cao, Dahua Lin, and Jiaqi Wang.
\newblock Dispider: Enabling video llms with active real-time interaction via disentangled perception, decision, and reaction.
\newblock \emph{arXiv preprint arXiv:2501.03218}, 2025.

\bibitem[Rahman et~al.(2024)Rahman, Shihab, Chu, and Sharma]{rahman2024deeplocalization}
Mohammed~Shaiqur Rahman, Ibne~Farabi Shihab, Lynna Chu, and Anuj Sharma.
\newblock Deeplocalization: Using change point detection for temporal action localization.
\newblock In \emph{Proceedings of the IEEE/CVF Conference on Computer Vision and Pattern Recognition}, pages 7252--7260, 2024.

\bibitem[Shen et~al.(2024)Shen, Fu, Chen, Zhang, Li, Sun, Wu, Lin, and Ji]{shen2024aligning}
Yunhang Shen, Chaoyou Fu, Peixian Chen, Mengdan Zhang, Ke~Li, Xing Sun, Yunsheng Wu, Shaohui Lin, and Rongrong Ji.
\newblock Aligning and prompting everything all at once for universal visual perception.
\newblock In \emph{Proceedings of the IEEE/CVF Conference on Computer Vision and Pattern Recognition}, pages 13193--13203, 2024.

\bibitem[Song et~al.(2024)Song, Chai, Wang, Zhang, Zhou, Wu, Chi, Guo, Ye, Zhang, Lu, Hwang, and Wang]{MovieChat}
Enxin Song, Wenhao Chai, Guanhong Wang, Yucheng Zhang, Haoyang Zhou, Feiyang Wu, Haozhe Chi, Xun Guo, Tian Ye, Yanting Zhang, Yan Lu, Jenq-Neng Hwang, and Gaoang Wang.
\newblock Moviechat: From dense token to sparse memory for long video understanding.
\newblock In \emph{2024 IEEE/CVF Conference on Computer Vision and Pattern Recognition (CVPR)}, pages 18221--18232, 2024.
\newblock \doi{10.1109/CVPR52733.2024.01725}.

\bibitem[Stappen et~al.(2023)Stappen, Dillmann, Striegel, V{\"o}gel, Flores-Herr, and Schuller]{stappen2023integrating}
Lukas Stappen, Jeremy Dillmann, Serena Striegel, Hans-J{\"o}rg V{\"o}gel, Nicolas Flores-Herr, and Bj{\"o}rn~W Schuller.
\newblock Integrating generative artificial intelligence in intelligent vehicle systems.
\newblock In \emph{2023 IEEE 26th International Conference on Intelligent Transportation Systems (ITSC)}, pages 5790--5797. IEEE, 2023.

\bibitem[Sul et~al.(2023)Sul, Han, and Lee]{sul2023mr}
Jinhwan Sul, Jihoon Han, and Joonseok Lee.
\newblock Mr. hisum: A large-scale dataset for video highlight detection and summarization.
\newblock \emph{Advances in Neural Information Processing Systems}, 36:\penalty0 40542--40555, 2023.

\bibitem[Tang et~al.(2025)Tang, Bi, Xu, Song, Liang, Wang, Zhang, An, Lin, Zhu, et~al.]{tang2025video}
Yunlong Tang, Jing Bi, Siting Xu, Luchuan Song, Susan Liang, Teng Wang, Daoan Zhang, Jie An, Jingyang Lin, Rongyi Zhu, et~al.
\newblock Video understanding with large language models: A survey.
\newblock \emph{IEEE Transactions on Circuits and Systems for Video Technology}, 2025.

\bibitem[Team et~al.(2023)Team, Anil, Borgeaud, Alayrac, Yu, Soricut, Schalkwyk, Dai, Hauth, Millican, et~al.]{team2023gemini}
Gemini Team, Rohan Anil, Sebastian Borgeaud, Jean-Baptiste Alayrac, Jiahui Yu, Radu Soricut, Johan Schalkwyk, Andrew~M Dai, Anja Hauth, Katie Millican, et~al.
\newblock Gemini: a family of highly capable multimodal models.
\newblock \emph{arXiv preprint arXiv:2312.11805}, 2023.

\bibitem[Team et~al.(2024)Team, Mesnard, Hardin, Dadashi, Bhupatiraju, Pathak, Sifre, Rivi{\`e}re, Kale, Love, et~al.]{team2024gemma}
Gemma Team, Thomas Mesnard, Cassidy Hardin, Robert Dadashi, Surya Bhupatiraju, Shreya Pathak, Laurent Sifre, Morgane Rivi{\`e}re, Mihir~Sanjay Kale, Juliette Love, et~al.
\newblock Gemma: Open models based on gemini research and technology.
\newblock \emph{arXiv preprint arXiv:2403.08295}, 2024.

\bibitem[Touvron et~al.(2023)Touvron, Martin, Stone, Albert, Almahairi, Babaei, Bashlykov, Batra, Bhargava, Bhosale, et~al.]{touvron22023llama}
Hugo Touvron, Louis Martin, Kevin Stone, Peter Albert, Amjad Almahairi, Yasmine Babaei, Nikolay Bashlykov, Soumya Batra, Prajjwal Bhargava, Shruti Bhosale, et~al.
\newblock Llama 2: Open foundation and fine-tuned chat models.
\newblock \emph{arXiv preprint arXiv:2307.09288}, 2023.

\bibitem[Wang et~al.(2023)Wang, Chen, Luo, Dai, Yuan, Wu, and Jiang]{wang2023chatvideotrackletcentricmultimodalversatile}
Junke Wang, Dongdong Chen, Chong Luo, Xiyang Dai, Lu~Yuan, Zuxuan Wu, and Yu-Gang Jiang.
\newblock Chatvideo: A tracklet-centric multimodal and versatile video understanding system, 2023.
\newblock URL \url{https://arxiv.org/abs/2304.14407}.

\bibitem[Wang et~al.(2024{\natexlab{a}})Wang, Bai, Tan, Wang, Fan, Bai, Chen, Liu, Wang, Ge, et~al.]{wang2024qwen2}
Peng Wang, Shuai Bai, Sinan Tan, Shijie Wang, Zhihao Fan, Jinze Bai, Keqin Chen, Xuejing Liu, Jialin Wang, Wenbin Ge, et~al.
\newblock Qwen2-vl: Enhancing vision-language model's perception of the world at any resolution.
\newblock \emph{arXiv preprint arXiv:2409.12191}, 2024{\natexlab{a}}.

\bibitem[Wang et~al.(2024{\natexlab{b}})Wang, Zhang, Zohar, and Yeung-Levy]{wang2024videoagent}
Xiaohan Wang, Yuhui Zhang, Orr Zohar, and Serena Yeung-Levy.
\newblock Videoagent: Long-form video understanding with large language model as agent.
\newblock In \emph{European Conference on Computer Vision}, pages 58--76. Springer, 2024{\natexlab{b}}.

\bibitem[Wang et~al.(2025)Wang, Zhang, Zohar, and Yeung-Levy]{wang2025videoagent}
Xiaohan Wang, Yuhui Zhang, Orr Zohar, and Serena Yeung-Levy.
\newblock Videoagent: Long-form video understanding with large language model as agent.
\newblock In \emph{European Conference on Computer Vision}, pages 58--76. Springer, 2025.

\bibitem[Wang et~al.(2024{\natexlab{c}})Wang, Yang, and Ren]{wang2024lifelongmemoryleveragingllmsanswering}
Ying Wang, Yanlai Yang, and Mengye Ren.
\newblock Lifelongmemory: Leveraging llms for answering queries in long-form egocentric videos, 2024{\natexlab{c}}.
\newblock URL \url{https://arxiv.org/abs/2312.05269}.

\bibitem[Wang et~al.(2024{\natexlab{d}})Wang, Xie, Liu, and Zheng]{wang2024videollamblongcontextvideounderstanding}
Yuxuan Wang, Cihang Xie, Yang Liu, and Zilong Zheng.
\newblock Videollamb: Long-context video understanding with recurrent memory bridges, 2024{\natexlab{d}}.
\newblock URL \url{https://arxiv.org/abs/2409.01071}.

\bibitem[Wei et~al.(2022)Wei, Wang, Schuurmans, Bosma, Xia, Chi, Le, Zhou, et~al.]{wei2022chain}
Jason Wei, Xuezhi Wang, Dale Schuurmans, Maarten Bosma, Fei Xia, Ed~Chi, Quoc~V Le, Denny Zhou, et~al.
\newblock Chain-of-thought prompting elicits reasoning in large language models.
\newblock \emph{Advances in neural information processing systems}, 35:\penalty0 24824--24837, 2022.

\bibitem[Weng et~al.(2024)Weng, Han, He, Chang, and Zhuang]{weng2024longvlm}
Yuetian Weng, Mingfei Han, Haoyu He, Xiaojun Chang, and Bohan Zhuang.
\newblock Longvlm: Efficient long video understanding via large language models.
\newblock In \emph{European Conference on Computer Vision}, pages 453--470. Springer, 2024.

\bibitem[Wu et~al.(2024)Wu, Chen, Lin, Wang, Gao, Xu, Xu, Hu, Chen, and Shou]{wu2024videollm}
Shiwei Wu, Joya Chen, Kevin~Qinghong Lin, Qimeng Wang, Yan Gao, Qianli Xu, Tong Xu, Yao Hu, Enhong Chen, and Mike~Zheng Shou.
\newblock Videollm-mod: Efficient video-language streaming with mixture-of-depths vision computation.
\newblock \emph{Advances in Neural Information Processing Systems}, 37:\penalty0 109922--109947, 2024.

\bibitem[Yang et~al.(2024{\natexlab{a}})Yang, Chen, Yu, Shen, and Gan]{yang2024vca}
Zeyuan Yang, Delin Chen, Xueyang Yu, Maohao Shen, and Chuang Gan.
\newblock Vca: Video curious agent for long video understanding.
\newblock \emph{arXiv preprint arXiv:2412.10471}, 2024{\natexlab{a}}.

\bibitem[Yang et~al.(2025)Yang, Hu, Du, Xue, Qian, Wu, Yang, Dong, and Xu]{yang2025svbench}
Zhenyu Yang, Yuhang Hu, Zemin Du, Dizhan Xue, Shengsheng Qian, Jiahong Wu, Fan Yang, Weiming Dong, and Changsheng Xu.
\newblock Svbench: A benchmark with temporal multi-turn dialogues for streaming video understanding.
\newblock \emph{arXiv preprint arXiv:2502.10810}, 2025.

\bibitem[Yang et~al.(2024{\natexlab{b}})Yang, Chen, Li, Wang, and Yang]{yang2024doraemongptunderstandingdynamicscenes}
Zongxin Yang, Guikun Chen, Xiaodi Li, Wenguan Wang, and Yi~Yang.
\newblock Doraemongpt: Toward understanding dynamic scenes with large language models (exemplified as a video agent), 2024{\natexlab{b}}.
\newblock URL \url{https://arxiv.org/abs/2401.08392}.

\bibitem[Yao et~al.(2024)Yao, Yu, Zhang, Wang, Cui, Zhu, Cai, Li, Zhao, He, et~al.]{yao2024minicpm}
Yuan Yao, Tianyu Yu, Ao~Zhang, Chongyi Wang, Junbo Cui, Hongji Zhu, Tianchi Cai, Haoyu Li, Weilin Zhao, Zhihui He, et~al.
\newblock Minicpm-v: A gpt-4v level mllm on your phone.
\newblock \emph{arXiv preprint arXiv:2408.01800}, 2024.

\bibitem[Zang et~al.(2023)Zang, Wang, Pei, and Liang]{zang2023discovering}
Chuanqi Zang, Hanqing Wang, Mingtao Pei, and Wei Liang.
\newblock Discovering the real association: Multimodal causal reasoning in video question answering.
\newblock In \emph{Proceedings of the IEEE/CVF Conference on Computer Vision and Pattern Recognition}, pages 19027--19036, 2023.

\bibitem[Zeng et~al.()Zeng, Li, Wang, Li, Jiang, Yan, Li, Shi, Yue, Wang, et~al.]{zengtimesuite}
Xiangyu Zeng, Kunchang Li, Chenting Wang, Xinhao Li, Tianxiang Jiang, Ziang Yan, Songze Li, Yansong Shi, Zhengrong Yue, Yi~Wang, et~al.
\newblock Timesuite: Improving mllms for long video understanding via grounded tuning.
\newblock In \emph{The Thirteenth International Conference on Learning Representations}.

\bibitem[Zhang et~al.(2024{\natexlab{a}})Zhang, Lin, Yang, Wang, Li, Lin, Liu, and Wang]{zhang2024mm}
Chaoyi Zhang, Kevin Lin, Zhengyuan Yang, Jianfeng Wang, Linjie Li, Chung-Ching Lin, Zicheng Liu, and Lijuan Wang.
\newblock Mm-narrator: Narrating long-form videos with multimodal in-context learning.
\newblock In \emph{Proceedings of the IEEE/CVF Conference on Computer Vision and Pattern Recognition}, pages 13647--13657, 2024{\natexlab{a}}.

\bibitem[Zhang et~al.(2024{\natexlab{b}})Zhang, Zhang, Zhang, and Tresp]{Zhang_2024_WACV}
Gengyuan Zhang, Yurui Zhang, Kerui Zhang, and Volker Tresp.
\newblock Can vision-language models be a good guesser? exploring vlms for times and location reasoning.
\newblock In \emph{Proceedings of the IEEE/CVF Winter Conference on Applications of Computer Vision (WACV)}, pages 636--645, January 2024{\natexlab{b}}.

\bibitem[Zhang et~al.(2023)Zhang, Li, and Bing]{zhang2023video}
Hang Zhang, Xin Li, and Lidong Bing.
\newblock Video-llama: An instruction-tuned audio-visual language model for video understanding.
\newblock In \emph{Proceedings of the 2023 Conference on Empirical Methods in Natural Language Processing: System Demonstrations}, pages 543--553, 2023.

\bibitem[Zhang et~al.(2024{\natexlab{c}})Zhang, Wang, Tang, Liu, Feng, Dai, and Jin]{zhang2024flashvstreammemorybasedrealtimeunderstanding}
Haoji Zhang, Yiqin Wang, Yansong Tang, Yong Liu, Jiashi Feng, Jifeng Dai, and Xiaojie Jin.
\newblock Flash-vstream: Memory-based real-time understanding for long video streams, 2024{\natexlab{c}}.
\newblock URL \url{https://arxiv.org/abs/2406.08085}.

\bibitem[Zhang et~al.(2024{\natexlab{d}})Zhang, Zhao, Ying, Ma, and Lee]{zhang2024omagentmultimodalagentframework}
Lu~Zhang, Tiancheng Zhao, Heting Ying, Yibo Ma, and Kyusong Lee.
\newblock Omagent: A multi-modal agent framework for complex video understanding with task divide-and-conquer, 2024{\natexlab{d}}.
\newblock URL \url{https://arxiv.org/abs/2406.16620}.

\bibitem[Zhang et~al.(2024{\natexlab{e}})Zhang, Li, Liu, Lee, Gui, Fu, Feng, Liu, and Li]{zhang2024llavanextvideo}
Yuanhan Zhang, Bo~Li, haotian Liu, Yong~jae Lee, Liangke Gui, Di~Fu, Jiashi Feng, Ziwei Liu, and Chunyuan Li.
\newblock Llava-next: A strong zero-shot video understanding model, April 2024{\natexlab{e}}.
\newblock URL \url{https://llava-vl.github.io/blog/2024-04-30-llava-next-video/}.

\bibitem[Zheng et~al.(2024)Zheng, Gou, Kil, Sun, and Su]{zheng2024gpt}
Boyuan Zheng, Boyu Gou, Jihyung Kil, Huan Sun, and Yu~Su.
\newblock Gpt-4v (ision) is a generalist web agent, if grounded.
\newblock In \emph{Proceedings of the 41st International Conference on Machine Learning}, pages 61349--61385, 2024.

\bibitem[Zhu et~al.(2023)Zhu, Chen, Shen, Li, and Elhoseiny]{zhu2023minigpt}
Deyao Zhu, Jun Chen, Xiaoqian Shen, Xiang Li, and Mohamed Elhoseiny.
\newblock Minigpt-4: Enhancing vision-language understanding with advanced large language models.
\newblock \emph{arXiv preprint arXiv:2304.10592}, 2023.

\bibitem[Zhukov et~al.(2019)Zhukov, Alayrac, Cinbis, Fouhey, Laptev, and Sivic]{CrossTask}
Dimitri Zhukov, Jean-Baptiste Alayrac, Ramazan~Gokberk Cinbis, David Fouhey, Ivan Laptev, and Josef Sivic.
\newblock Cross-task weakly supervised learning from instructional videos.
\newblock In \emph{2019 IEEE/CVF Conference on Computer Vision and Pattern Recognition (CVPR)}, pages 3532--3540, 2019.
\newblock \doi{10.1109/CVPR.2019.00365}.

\end{thebibliography}
}

\newpage
\appendix

\section{Appendix}
In the Appendix, we include:
\begin{enumerate}
    \item Performance comparison with state-of-the-art Streaming Video baselines in Sec.~\ref{sec:more-ex}.
    \item More details of task definition (\eg query complexity, query temporality, and query category) in Sec .~\ref{sec:more-task};
    \item Details of \dataset construction and statistics in Sec.~\ref{sec:more-dataset};
    \item Implementation details including prompt engineering and hyperparameters we have used in \model in Sec.~\ref{sec:more-model};
    \item Extended experimental results including model performance on different query categories in Sec.~\ref{sec:more-exp2}.
\end{enumerate}

\subsection{Comaprison with SOTA Streaming Video Baselines} \label{sec:more-ex}
We compare our model compared to the state-of-the-art Streaming Video understanding models Flash-VStream~\cite{zhang2024flashvstreammemorybasedrealtimeunderstanding} and VideoLLM-Online~\cite{Chen_2024_CVPR}. As shown in Table~\ref{tab:otherwork}, \model sets a new standard by achieving the highest overall accuracy (61.5\%) and consistently outperforms both baselines across all temporal categories. Notably, it also demonstrates superior temporal grounding, achieving the lowest mean response delay (17.3s) and minimal variability (20.6s std), highlighting its reliability and temporal precision.


\begin{table}[!h]
\centering
\scriptsize
\setlength{\tabcolsep}{3pt} 
\caption{
Performance comparison with existing Streaming Video understanding models.}
\begin{tabular}{l|ccccc|cc}
\toprule
\multirow{2}{*}{\textbf{Trigger}} & \multicolumn{5}{c|}{\textbf{Accuracy (\%)}} & \multicolumn{2}{c}{\textbf{Temporal Offset (s)}} \\
 & \textbf{Overall}↑ & \pastq↑ & \presentq↑ & \futurehypoq↑ & \futurefactq↑ & \textbf{Mean $\delta$}$\downarrow$ & \textbf{Std $\delta$}$\downarrow$ \\
\midrule
VideoLLM-Online~\cite{Chen_2024_CVPR} & 9.6 & 5.8 & 12.0 & 11.8 & 9.3 & 69.7  & 257.5 \\
Flash-VStream~\cite{zhang2024flashvstreammemorybasedrealtimeunderstanding} & 36.2 & 33.1 & 38.2 & 35.1 & 38.3 & 53.1  & 221.3 \\
\textbf{\model} (ours) & \textbf{61.5} & \textbf{45.6} & \textbf{73.8} & \textbf{70.1} & \textbf{55.2} & \textbf{17.3} & \textbf{20.6} \\
\bottomrule
\end{tabular}
\label{tab:otherwork}
\end{table}


VideoLLM-Online is designed for streaming video understanding and assistant-style interaction. Users can query the model at any point during video playback. However, the model is optimized for shorter clips and struggles with our long-form video setting (sometimes exceeding 1 hour). 
This leads to a low overall accuracy (9.6\%) and considerable temporal offset variability (257.5s). 

Flash-VStream offers real-time QA during playback but lacks a mechanism to defer responses until sufficient context is available. This leads to premature and often imprecise answers, reflected in its modest accuracy (36.2\%) and high timing variance (std = 221.3s). Without temporal buffering or evidence-aware triggering, the model is unable to optimize when to respond.

In contrast, \model leverages evidence-grounded triggers and fine-grained temporal alignment to provide both accurate and timely responses, making it a robust choice for streaming video understanding at scale.

\subsection{Task Definition}\label{sec:more-task}
\label{sec:taxonomy}
We rigorously design our task considering the query temporality, complexity, and category. We have introduced the query temporality in Sec.~\ref{subsec:benchmark}.
We introduce the following taxonomy in \dataset considering  \textit{task complexity} and \textit{query category}.

\subsubsection{Task Complexity}
To better annotate \dataset and analyze agent performance, we categorize all queries into three levels of complexity as a general guideline: \textit{Perception}, \textit{Reasoning}, and \textit{Planning}.

\textbf{\textit{Perception}} tasks involve detecting and recognizing visual elements in the scene, such as objects, actions, spatial relationships \etc based on raw observations as images or video frames. The responses should be grounded in visual input from the past, present, or coming future yet do not require further reasoning.

\textbf{\textit{Reasoning}} tasks require the model to make inferences and reason, such as identifying causal relationships between events, or justifying the occurrence of events. This goes beyond surface-level perception to explain past events, interpret current event, and predict future events.

\textbf{\textit{Planning}} tasks demand a higher-level understanding of goal-directed behavior and temporal sequencing. These tasks involve anticipating future developments or strategizing future actions based on accumulated visual evidence and contextual cues.


\subsubsection{Query Category}
\begin{figure}[!h]
    \centering
    \includegraphics[width=1\linewidth]{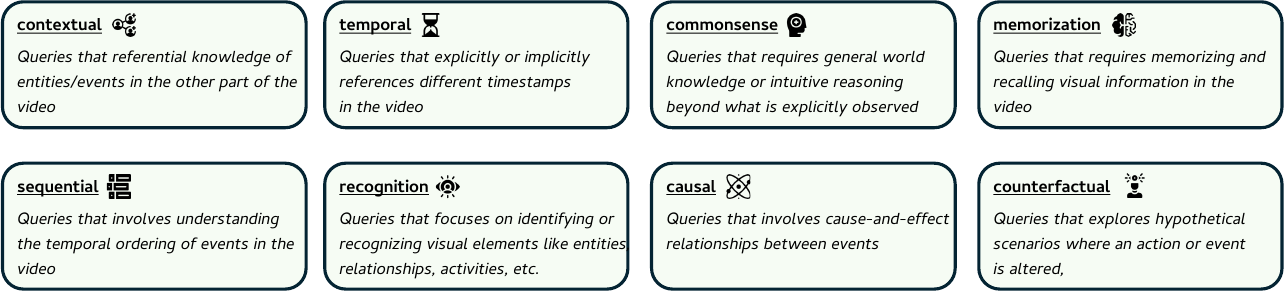}
    \caption{We have identified 8 different query categories considering the formulation and content of the query. This categorization is not mutually exclusive, which means a query could be tagged by the annotates as multiple categories following the guidelines.}
    \label{fig:qa_categories}
\end{figure}

To better understand how models perform under \qea, we define fine-grained query types. Different query categories demand different types of contextual and temporal knowledge from models to respond to the query. Evidence can be dispersed at different time points of the video.
As illustrated in Fig.~\ref{fig:qa_categories}, we define eight query categories, each characterized by its scope of reasoning. This categorization highlights the particular difficulty posed by the \qea setting, which introduces highly variable queries and demands dense memorization and contextual integration across the entire video, spanning both past and future events.
The query numbers of each category is shown in Tab.~\ref{fig:category-dist}.

\begin{itemize}
    \item \textbf{contextual} The question requires cross-referencing past events, entities, or objects that appeared earlier in the video.
    \item \textbf{temporal} The question explicitly or implicitly references different timestamps in the video (\eg, ``10 seconds ago,'' ``at 10s,'' ``before/when/after an event'').
    \item \textbf{commonsense} The question requires general world knowledge or intuitive reasoning beyond what is explicitly shown in the video. It may involve physical laws, social norms, or everyday experiences to infer answers.
    \item \textbf{memorization} The question requires recalling previously seen visual information from earlier in the video without relying on reasoning or commonsense knowledge. It tests the ability to retain and retrieve past observations.
    \item \textbf{sequential} The question involves understanding the order of discrete events in the video (before/after the current step).
    \item \textbf{recognition} The question focuses on identifying or describing entities, relationships, movements, scenes, summaries, or narration predictions.
    \item \textbf{causal} The question involves cause-and-effect relationships between events, requiring an understanding of how one action leads to another.
    \item \textbf{counterfactual} The question explores hypothetical scenarios where an action or event is altered, requiring reasoning about alternate possibilities.
\end{itemize}



Importantly, these categories are \textbf{not} mutually exclusive. As shown in Fig.~\ref{fig:question-example}, ``\texttt{Why does the camera holder open the washing machine}" can be both contextual and causal since it requires understanding the pre-context in the video and causal inference to justify the observation. This taxonomy enables a systematic evaluation of how well models manage different reasoning challenges in a temporally dynamic video context.

\subsubsection{Clarification of Temporality}
As shown in Tab.~\ref{tab:cla}, we aim to clarify the temporal relationship between evidence and query. For queries of different temporality, the evidence is entailed in different parts of the videos, which is coined as the asynchrony. For \pastq, the evidence can only happen before the query. While for the other three types, answering the questions could require knowledge about the current observation and/or observation in the past and future.  

\begin{table}[!h]
\centering
\footnotesize
\caption{Relationship of evidence and query in different query temporality.}
\label{tab:cla}
\begin{tabular}{lcccc}
\toprule
query category & \pastq & \presentq  & \futurehypoq & \futurefactq \\
\midrule
Evidence-\textit{before}-Query &  \checkmark  & \checkmark & \checkmark \\
Evidence-\textit{during}-Query  &         & \checkmark & \checkmark & \checkmark \\
Evidence-\textit{after}-Query  &         &         &       &   \checkmark  \\
\bottomrule
\end{tabular}
\end{table}

\subsection{Dataset: \dataset}
\label{sec:more-dataset}
In the following, we present details of the data construction and statistics of \dataset.

\subsubsection{Video Sources} Our proposed diagnostic dataset, \dataset, comprises a carefully curated selection of 189 videos drawn from existing datasets. 

\textbf{Ego4D} \cite{Ego4D} is a large-scale dataset of first-person (egocentric) videos capturing everyday human activities from a wide variety of environments, perspectives, and tasks. Recorded using head-mounted cameras, the video content emphasizes real-world, time-extended interactions and offers a rich source for modeling temporal and contextual understanding in egocentric vision tasks. In this work, only the raw video data is used, without leveraging the original annotations or task definitions. Ego4D is available under EGO4D License Agreement.

\textbf{MovieChat-1K}~\cite{MovieChat} provides a curated collection of movie video clips sourced from a diverse set of films, covering a wide range of scenes, characters, and narrative contexts. While originally designed for multimodal QA tasks, the dataset's primary value in this work lies in its high-quality video segments. We only used videos from the test set. MovieChat is available under BSD 3-Clause License.

\textbf{CrossTalk}~\cite{CrossTask} curates instructional videos in multiple handcraft domains such as making pancakes. It provides multi-grained events and strong temporal and causal relationships between events. This highlights a contextual understanding of the video content and we aim to adapt it into a more realistic streaming setting.


\textbf{Video-MME} \cite{Video-MME} is a full-spectrum, multi-modal evaluation benchmark for MLLMs. It was constructed to improve the long context modeling capabilities of MLLMs. In 900 raw videos from YouTube, it covers 30 fine-grained categories like astronomy, esports, or magic show. As the videos are grouped according to their length, we focused on \textit{long videos} ranging 30 to 60 minutes here. Video-MME is available under CC BY-SA 4.0.

\begin{figure}[htbp]
    \centering
    \begin{subfigure}[t]{0.52\textwidth}
        \centering
        \includegraphics[width=\linewidth]{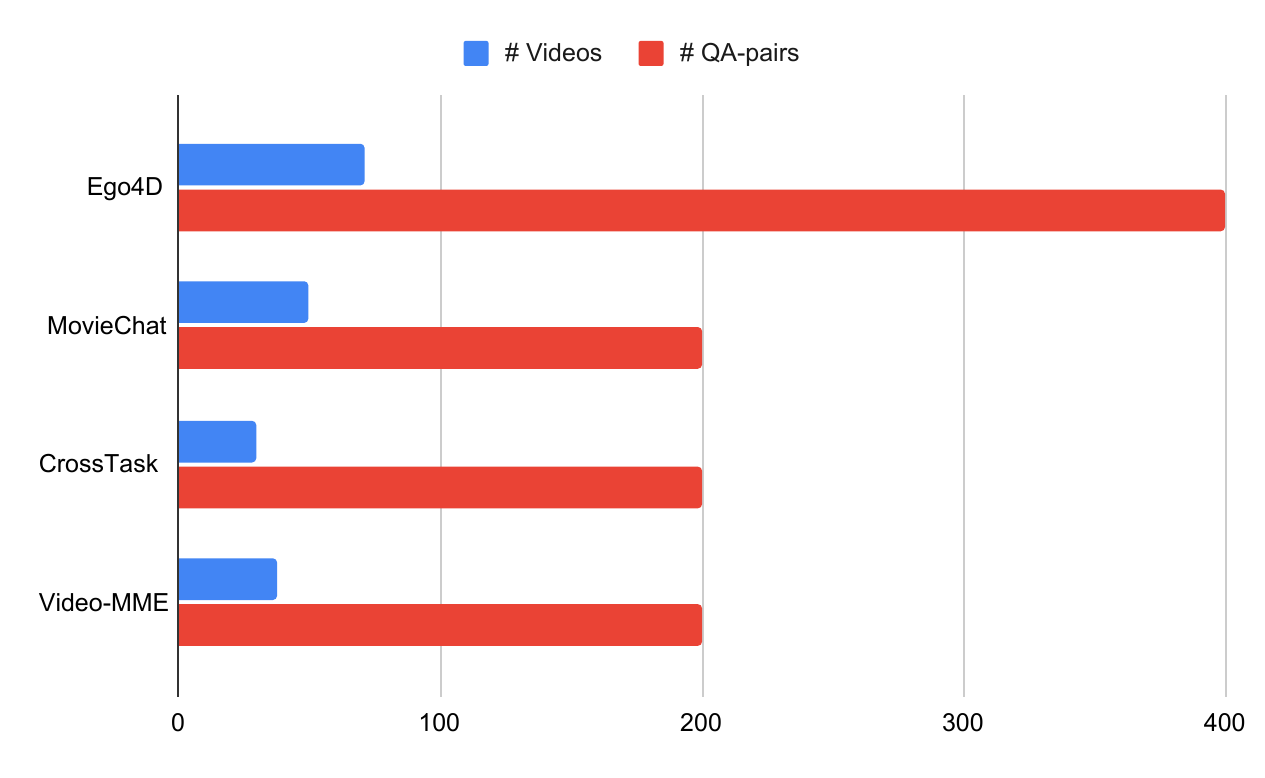}
        \caption{We selected 189 videos from four different datasets. The horizontal bars represent the number of videos (blue) and QA pairs (red) for each dataset: Ego4D, MovieChat, CrossTask, and Video-MME.}
        \label{fig:distr_data}
    \end{subfigure}
    \hfill
    \begin{subfigure}[t]{0.4\textwidth}
        \centering
        \includegraphics[width=\linewidth]{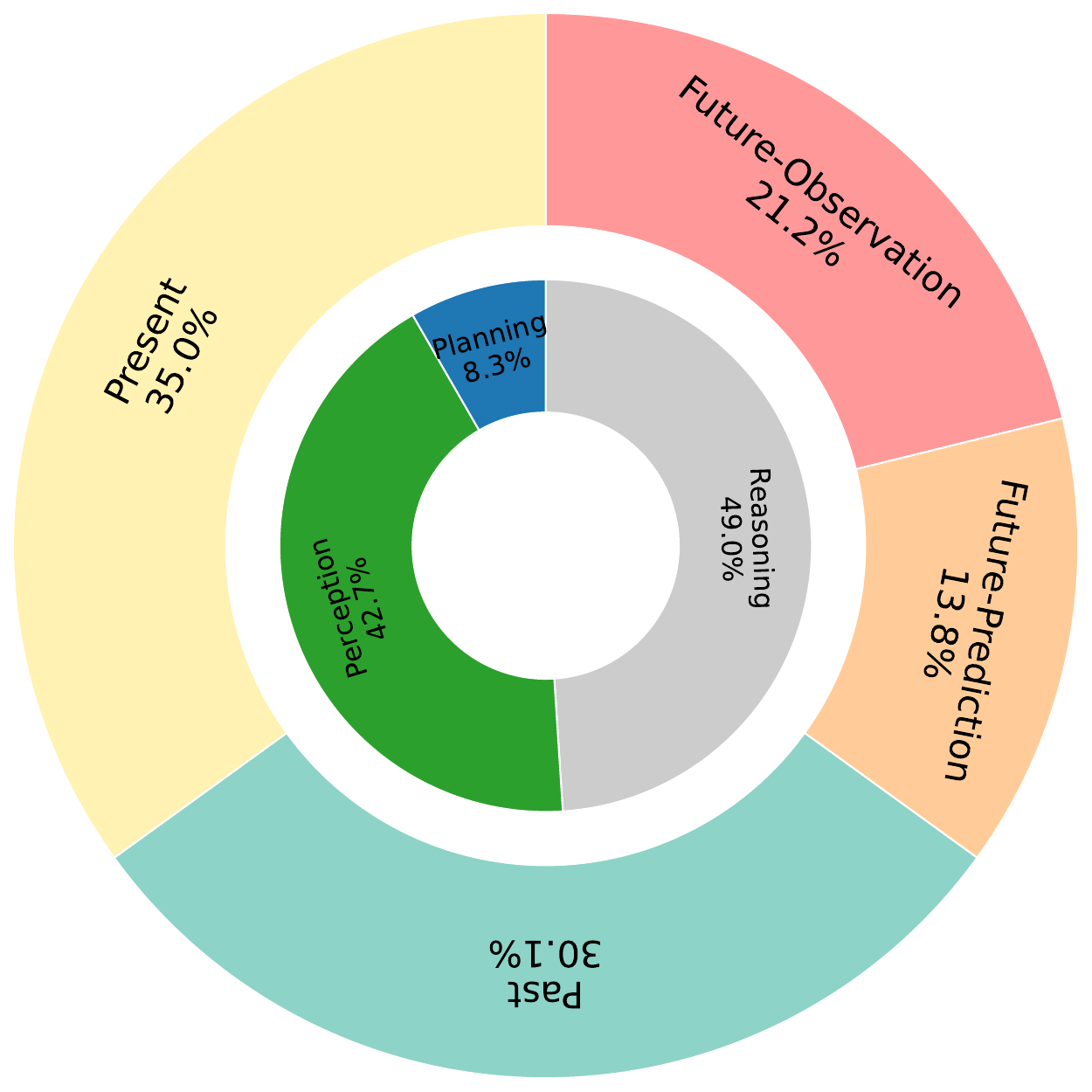}
        \caption{Query distribution categorized by task complexity (inner ring) and query temporality (outer ring).}
        \label{fig:nested_pie_chart}
    \end{subfigure}

    \vspace{0.8em}

    \begin{subfigure}[t]{0.6\textwidth}
        \centering
        \includegraphics[width=\linewidth]{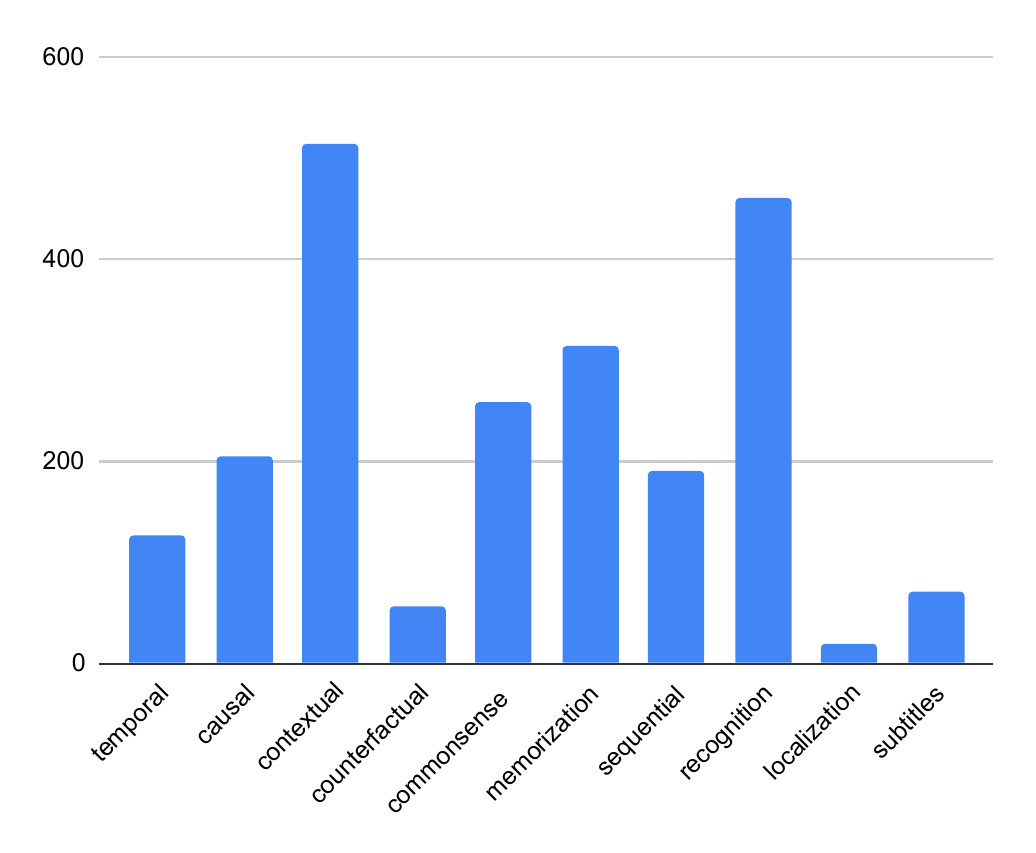}
        \caption{Number of questions according to query categories.}
        \label{fig:category-dist}
    \end{subfigure}

    \caption{Overview of data and question type distributions across multiple dimensions.}
    \label{fig:combined}
\end{figure}

The distribution of different data sources is shown in Fig.~\ref{fig:distr_data}.

\subsubsection{License}
We use standard licenses from the community
and provide the following links to the licenses for the datasets that we used in this paper. 


\noindent\textbf{Ego4D\cite{Ego4D}:} \href{https://github.com/facebookresearch/Ego4d/blob/main/LICENSE}{MIT}

\noindent\textbf{MovieChat-1K~\cite{MovieChat}:} \href{https://github.com/rese1f/MovieChat/blob/main/LICENSE}{BSD 3-Clause}

\noindent\textbf{CrossTalk~\cite{CrossTask}:} \href{https://github.com/DmZhukov/CrossTask?tab=BSD-3-Clause-1-ov-file}{BSD 3-Clause}

\noindent\textbf{Video-MME~\cite{Video-MME}:} \href{https://github.com/MME-Benchmarks/Video-MME?tab=readme-ov-file#-dataset}{This dataset is released for academic research only. Commercial use is prohibited. Redistribution, modification, and dissemination are not allowed without prior approval from the authors.}

\subsubsection{Video Selection} To construct a challenging benchmark that demands strong reasoning capabilities and temporal awareness, we selected videos ranging from 6 to 90 minutes in length across all four datasets (see Fig.~\ref{fig:video_len}). Our selection emphasizes diversity in both content and perspective: videos span a wide array of domains, including cooking, driving, construction, mechanics, astronomy, nature, animals, television shows, and everyday life. We intentionally include both videos of egocentric and third-person perspectives to encourage robust generalization across different viewpoints in a realistic streaming.

\subsubsection{Dataset Construction}
\textbf{Annotation} The dataset comprises 1,000 carefully curated question-answer pairs, annotated and reviewed over the course of more than 400 hours by five independent annotators. 
Annotators select videos from the provided datasets following the selection criteria described above. For each chosen video, the annotators are guided to watch the whole video sequentially without pause and compose complex and challenging questions spanning various task complexities and query temporalities, following our annotation guidelines.

To identify correct answers, annotators are requested to watch each video multiple times after question generation, ensuring accuracy for the ground-truth answers and designing realistic yet incorrect candidate answers for the multiple-choice question answering setting. 
Each answer is also paired with a \textit{proposed answering window} annotated by the same annotator to ground the answer in the video temporally.
Each question-answer pair was also tagged with its relevant query categories (see \ref{sec:taxonomy}) according to the query content.

\textbf{Reviewing} To maintain high annotation quality, a rigorous review process was implemented. Each question is reviewed by a reviewer other than the annotator, including answer validity, proposed answering window, and query categories. 
Besides, all QA pairs were thoroughly checked for spelling, clarity, relevance, difficulty, and the quality of both questions and answer choices.

For the Ego4D videos, we additionally incorporated questions generated by ChatGPT~\footnote{https://chatgpt.com/} with the rich annotations in the original dataset and the prompt used in ~\cite{nagrani2024neptune}, which represent 10.8\% of the entire dataset. These automatically generated questions were subsequently reviewed and refined by a human annotator to meet high-quality standards and to ensure accuracy.

\subsubsection{Statistics}

To comprehensively reflect the temporal dynamics of streaming video interactions, the QA pairs are uniformly distributed across past, present, and future contexts, with the future category further divided into \futurefactq and \futurehypoq questions, as shown in Fig.~\ref{fig:nested_pie_chart}.

Crucially, \dataset emphasizes not just perception-based queries but also reasoning involving commonsense knowledge and contextual understanding. Approximately half of the questions require reasoning (see Fig.~\ref{fig:nested_pie_chart}), encompassing temporal, causal, commonsense, counterfactual, and planning-based inference. 

The benchmark also captures a broad spectrum of query categories, including temporal, causal, contextual, counterfactual, commonsense, memorization, sequential, recognition, localization, and subtitle-based questions. This diversity enables robust and multifaceted evaluation of model capabilities across understanding, reasoning, and temporal alignment in real-time video streams.

\subsection{Implementation Details}\label{sec:more-model}
We elucidate on the details of \model in this section for reproducibility.

\subsubsection{Ingestion Tool Use}
We employ a modular ingestion pipeline composed of the following components, each responsible for extracting structured information from raw video inputs and populating our memory bank:

\noindent \textbf{Multimodal Large Language Models for Captioning:}  
We use the following models for captioning: \textit{LLaVA‑NeXT‑Video}~\cite{zhang2024llavanextvideo}, \textit{LLaVA‑OneVision}~\cite{li2024llavaonevisioneasyvisualtask}, \textit{Qwen2.5‑VL}~\cite{bai2025qwen2}, and \textit{Qwen2‑VL‑Instruct}~\cite{wang2024qwen2}.  
These models process each video frame or segment to generate dense and temporally grounded captions, enabling downstream reasoning about actions, objects, and events across time.

\vspace{0.5em}
\noindent \textbf{Object Detection Module:}  
Leveraging the state-of-the-art detector from~\cite{shen2024aligning}, we identify and localize key objects in each frame. Detected bounding boxes are embedded alongside captions, which facilitates fine-grained object-centric memory queries.

\vspace{0.5em}
\noindent \textbf{Rule-Based Scene Graph Generator:}  
Using syntactic and semantic parsing over detection outputs, this component builds structured scene graphs. Entities are linked by relations such as \textit{subject–verb–object} and spatial predicates, helping represent visual scenes in a relational format.

\vspace{0.5em}
\noindent \textbf{Memory Indexing via Vector Database:}  
We embed captions and video frames into continuous representations and store them using a scalable vector store (e.g., FAISS~\cite{johnson2017billion}). This enables efficient similarity-based retrieval for contextual grounding during question answering, with support for nearest-neighbor lookups.

\subsubsection{Prompt Engineering}

We design specialized prompts to guide different stages of the video understanding pipeline, including caption generation, ingestion validation, and answer reasoning.

\vspace{0.5em}
\noindent \textbf{Captioning Prompt:}
\begin{quote}
\texttt{
You are a video captioning assistant. Your task is to generate a detailed caption describing what is happening in the video. Focus on describing the key actions, objects, and events in the video. Be specific and detailed in your description.
}
\end{quote}

\noindent \textbf{Binary QA trigger Prompt:}
\begin{quote}
\texttt{
You are a video understanding assistant. Your task is to determine if the given question \{CAN\} be answered based on the video content and the memory context.\\
Respond with only \texttt{true} if the question can be answered, or \texttt{false} if it cannot be answered based on the video content and the memory bank.\\
Do not provide any explanation, only respond with \texttt{true} or \texttt{false}.
}
\end{quote}

\noindent \textbf{CoT Evidence Reasoning Prompt:} 
\begin{quote}
\texttt{You are a video understanding assistant. Your task is to determine if the given question can be answered based on the video content and the memory context.\\
Let's think step by step to assess the available evidence. Conclude with a single \texttt{true} or \texttt{false} based on whether the question can be answered.\\
End your response with only \texttt{true} or \texttt{false}.}
\end{quote}

\noindent \textbf{Adversarial Verification Rejection Prompt:}
\begin{quote}
\texttt{
You are a video understanding assistant. Your task is to determine if the given question \textbf{CANNOT} be answered based on the video content and the memory context.\\
Respond with only \texttt{true} if the question should be rejected (i.e., cannot be answered), or \texttt{false} if it should not be rejected.\\
Do not provide any explanation, only respond with \texttt{true} or \texttt{false}.
}
\end{quote}

\vspace{0.5em}
\noindent \textbf{Reasoning Prompt:}
\begin{quote}
\texttt{
You are a video reasoning assistant. Your task is to answer questions about the video content. Use the provided context and video content to formulate your answer. If you are uncertain about any aspect of the answer, acknowledge the uncertainty. Only respond with the candidate's answer label.
}
\end{quote}

\begin{figure*}
    \centering
    \begin{subfigure}[b]{0.32\linewidth}
        \centering
        \includegraphics[width=\linewidth]{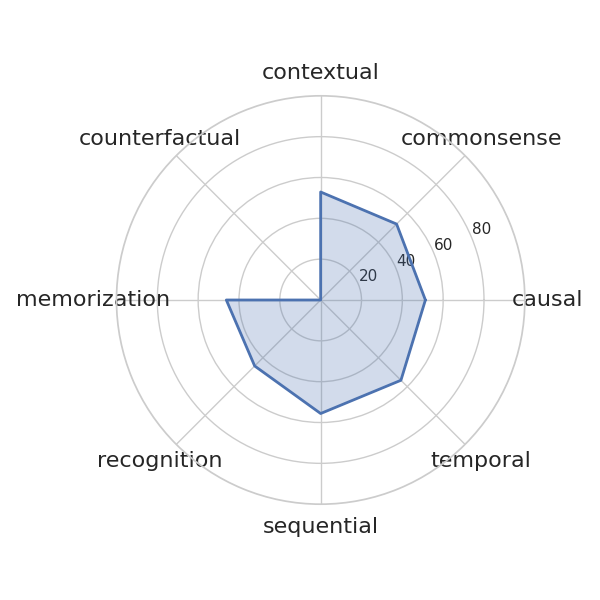}
        \caption{\scriptsize Qwen2-VL-7B-Instruct with Text Memory and Binary QA Trigger}
    \end{subfigure}
    \hfill
    \begin{subfigure}[b]{0.32\linewidth}
        \centering
        \includegraphics[width=\linewidth]{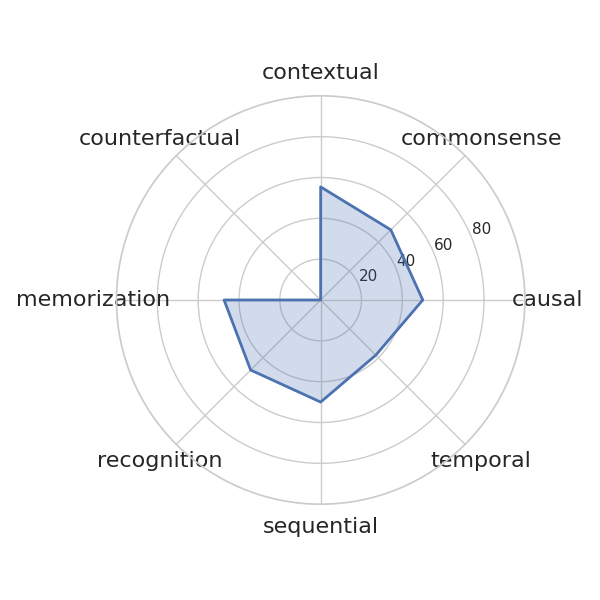}
        \caption{\scriptsize Qwen2-VL-7B-Instructtext with Vision Memory and Binary QA Trigger}
    \end{subfigure}

    \vspace{0.3em}
    \begin{subfigure}[b]{0.32\linewidth}
        \centering
        \includegraphics[width=\linewidth]{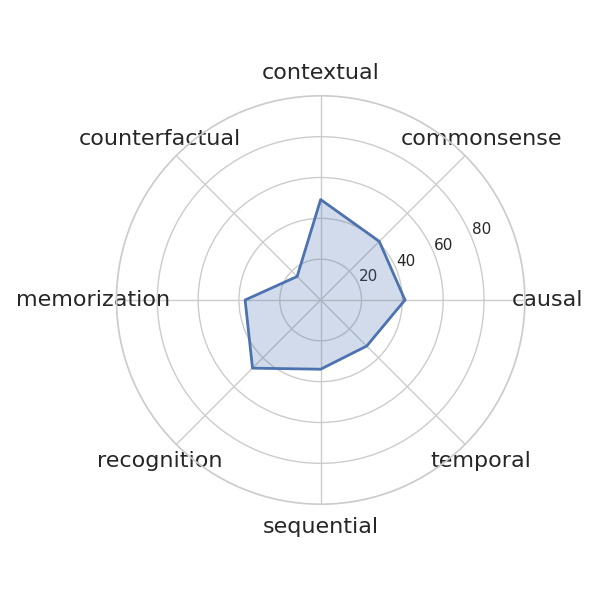}
        \caption{\scriptsize Qwen2-VL-7B-Instruct with Vision Memory and Binary QA Trigger}
    \end{subfigure}    
    \hfill
    \begin{subfigure}[b]{0.32\linewidth}
        \centering
        \includegraphics[width=\linewidth]{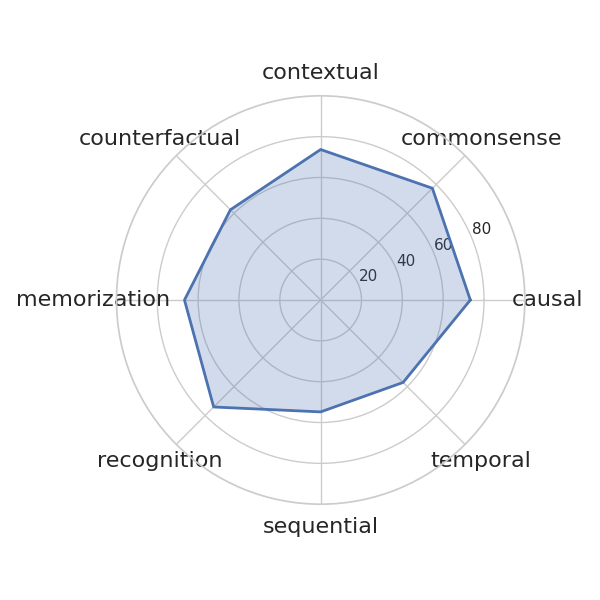}
        \caption{\scriptsize Qwen2-VL-7B-Instruct with Text Memory and Adversarial Verification Trigger}
    \end{subfigure}

    \vspace{0.3em}
    \begin{subfigure}[b]{0.32\linewidth}
        \centering
        \includegraphics[width=\linewidth]{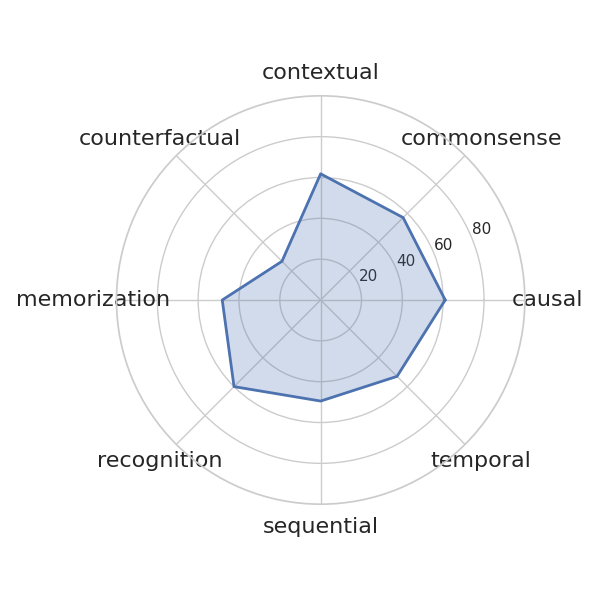}
        \caption{\scriptsize Qwen2-VL-7B-Instruct with text memory and CoT Reasoning Trigger}
    \end{subfigure}
    \hfill
    \begin{subfigure}[b]{0.32\linewidth}
        \centering
        \includegraphics[width=\linewidth]{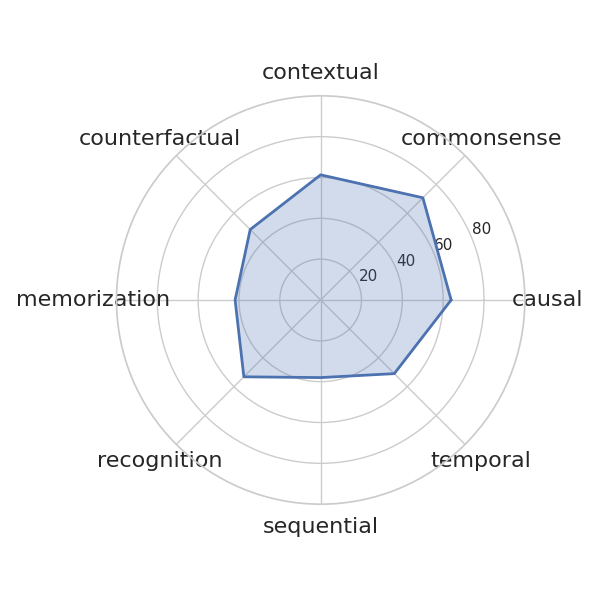}
        \caption{\scriptsize Qwen2.5-VL-3B-Instruct with Text memory and CoT Reasoning Trigger}
    \end{subfigure}

    \vspace{0.3em}
    \begin{subfigure}[b]{0.32\linewidth}
        \centering
        \includegraphics[width=\linewidth]{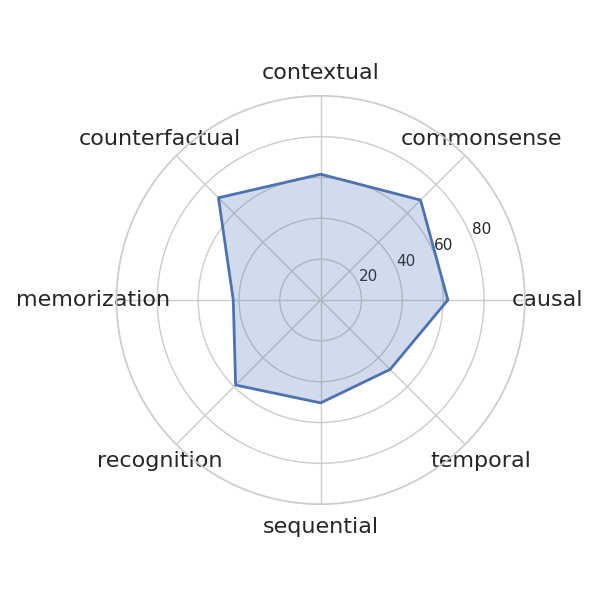}
        \caption{\scriptsize LLava-OneVision-qwen2-7b-ov-hf with Text Memory and CoT Reasoning Trigger}
    \end{subfigure}
    \hfill
    \begin{subfigure}[b]{0.32\linewidth}
        \centering
        \includegraphics[width=\linewidth]{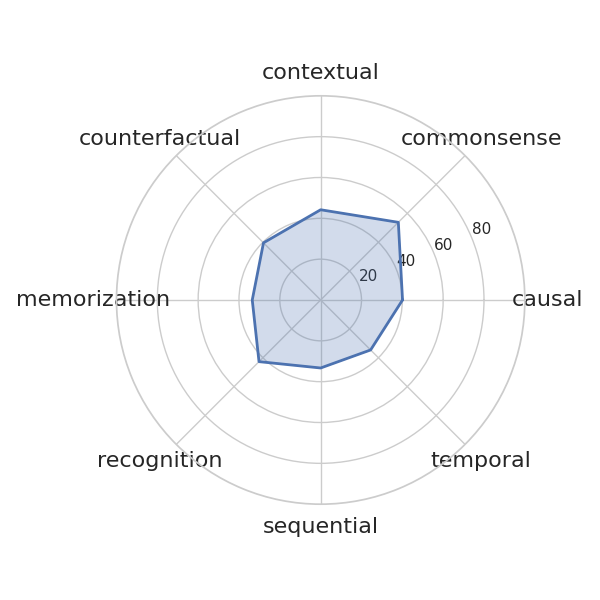}
        \caption{\scriptsize LLaVA-NeXT-Video-7B-hf and Text Memory and CoT Reasoning Trigger}
    \end{subfigure}
    \caption{\small Performance comparison across different models and reasoning types.}
    \label{fig:combined_radar_compact}
\end{figure*}

\subsection{Performance on Different Task Categories} ~\label{sec:more-exp2}
\begin{figure*}
    \begin{subfigure}[b]{0.4\linewidth}
        \centering
        \includegraphics[width=\linewidth]{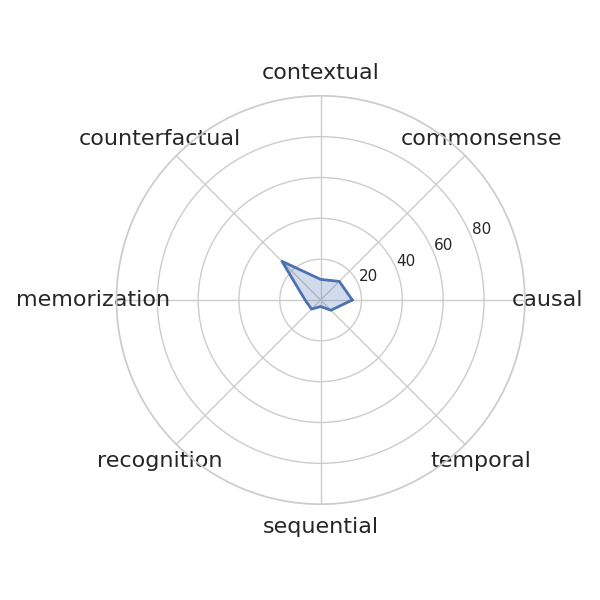}
        \caption{\scriptsize VideoLLM-Online }
    \end{subfigure}
    \hfill
    \begin{subfigure}[b]{0.4\linewidth}
        \centering
        \includegraphics[width=\linewidth]{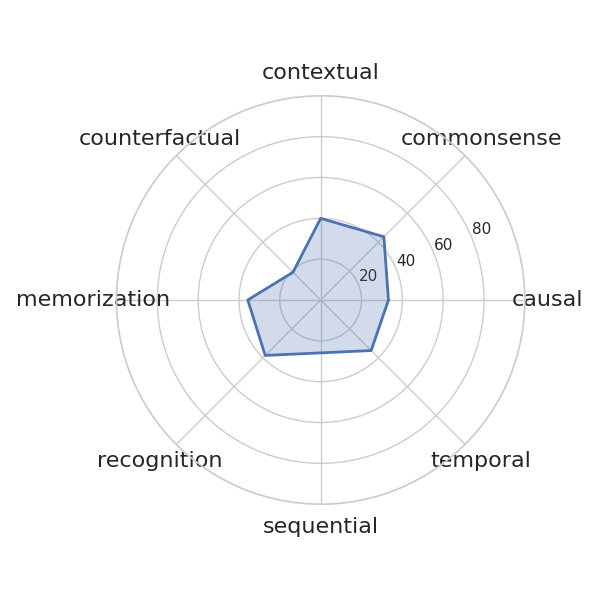}
        \caption{\scriptsize Flash-Vstream}
    \end{subfigure}
    \caption{\small Performance comparisons across different query types with SOTA streaming video models.}
    \label{fig:combined_radar_compact2}
\end{figure*}

We provide radar charts in Fig~\ref{fig:combined_radar_compact} and \ref{fig:combined_radar_compact2} to illustrate model performance across eight complexity dimensions under various prompting triggers and SOTA streaming video understanding models. Binary QA Trigger achieves moderate performance with noticeable limitations in temporal and counterfactual tasks. CoT Reasoning Trigger enhance memorization but at the cost of temporal coherence. Adversarial Verification Trigger consistently improves causal, temporal, and contextual reasoning. As shown in Fig.~\ref{fig:combined_radar_compact2}, the previous streaming video models achieve subpar performance across all the task categories compared to our proposed model \model.

\end{document}